\documentclass[10pt,twocolumn,letterpaper]{article}

\usepackage[pagenumbers]{cvpr} % To force page numbers, e.g. for an arXiv version

\usepackage{graphicx}
\usepackage{amsmath}
\usepackage{amssymb}
\usepackage{booktabs}
\usepackage{multirow}
\usepackage{makecell}
\usepackage{comment}
\usepackage{amsfonts}
\usepackage{epsfig}
\usepackage{diagbox}

\def\eg{\emph{e.g.}} 
\def\ie{\emph{i.e.}} 
 
\def\etc{\emph{etc.}} 
\def\etal{\emph{et al. }}
\newcommand{\BEST}[1]{\textbf{\textcolor[rgb]{1.00,0.00,0.00}{#1}}}
\newcommand{\SBEST}[1]{\textbf{\textcolor[rgb]{0.00,0.00,1.00}{#1}}}

\usepackage[pagebackref,breaklinks,colorlinks]{hyperref}

\usepackage[capitalize]{cleveref}
\crefname{section}{Sec.}{Secs.}
\Crefname{section}{Section}{Sections}
\Crefname{table}{Table}{Tables}
\crefname{table}{Tab.}{Tabs.}

\def\confName{CVPR}
\def\confYear{2022}

\begin{document}

%%%%%%%%% TITLE - PLEASE UPDATE
\title{EDTER: Edge Detection with Transformer}

\author{Mengyang Pu$^{1,3}$, Yaping Huang$^{1}$\thanks{Corresponding author.}, Yuming Liu$^{2}$, Qingji Guan$^{1}$, Haibin Ling$^{3}$\\ 
$^1$Beijing Key Laboratory of Traffic Data Analysis and Mining, Beijing Jiaotong University, China\\
$^2$Shenzhen Urban Transport Planning Center Co.,Ltd., China\\
$^3$Department of Computer Science, Stony Brook University, USA\\
{\tt\small \{mengyangpu, yphuang, qjguan\}@bjtu.edu.cn;\quad liuyuming@sutpc.com;\quad hling@cs.stonybrook.edu}
}
\maketitle

%%%%%%%%% ABSTRACT
\begin{abstract}
    Convolutional neural networks have made significant progresses in edge detection by progressively exploring the context and semantic features. However, local details are gradually suppressed with the enlarging of receptive fields. Recently, vision transformer has shown excellent capability in capturing long-range dependencies. Inspired by this, we propose a novel transformer-based edge detector, \emph{Edge Detection TransformER (EDTER)}, to extract clear and crisp object boundaries and meaningful edges by exploiting the full image context information and detailed local cues simultaneously. EDTER works in two stages. In Stage I, a global transformer encoder is used to capture long-range global context on coarse-grained image patches. Then in Stage II, a local transformer encoder works on fine-grained patches to excavate the short-range local cues. Each transformer encoder is followed by an elaborately designed Bi-directional Multi-Level Aggregation decoder to achieve high-resolution features. Finally, the global context and local cues are combined by a Feature Fusion Module and fed into a decision head for edge prediction. Extensive experiments on BSDS500, NYUDv2, and Multicue demonstrate the superiority of EDTER in comparison with state-of-the-arts. The source code is available at \url{https://github.com/MengyangPu/EDTER}.
\end{abstract}

%%%%%%%%% BODY TEXT
%---------------------------------------------------------------
\section{Introduction}
Edge detection is one of the most fundamental problems in computer vision and has a wide variety of applications, such as image segmentation~\cite{long2015fully,ronneberger2015u,chen2017deeplab,He2017MaskR,pinheiro2016learning,pinheiro2015learning}, object detection~\cite{He2017MaskR}, and video object segmentation~\cite{caelles2017one,voigtlaender2019feelvos,wang2019fast}. Given an input image, edge detection aims to extract accurate object boundaries and visually salient edges. It is challenging due to many factors including complex backgrounds, inconsistent annotations, and so on.

Edge detection is closely related to the context and semantic image cues. It is thus crucial to obtain appropriate representation to capture both high and low level visual cues. Traditional methods~\cite{kittler1983accuracy,canny1986computational,winnemoller2011xdog,martin2004learning,dollar2006supervised,lim2013sketch} mostly obtain edges based on low-level local cues, \eg, color and texture. Benefiting from the effectiveness of convolutional neural networks (CNNs) in learning semantic features, significant progress has been made for edge detection~\cite{kokkinos2015pushing,shen2015deepcontour,bertasius2015deepedge,bertasius2015hfl}. The CNN features progressively capture global and semantic-aware visual concepts with the enlargement of the receptive fields, while many essential fine details are inevitably and gradually lost at the mean time. To include more details, methods in~\cite{xie2015hed,liu2016rds,liu2017rcf,xu2017AMHNet,he2019bdcn} aggregate the features of deep and shallow layers. However, such shallow features reflect mainly local intensity variation without considering semantic context, resulting in noisy edges. 

\begin{figure}[!t]
\setlength{\abovecaptionskip}{3pt}
\setlength{\belowcaptionskip}{0pt}
\centering
\includegraphics[width=0.95\linewidth]{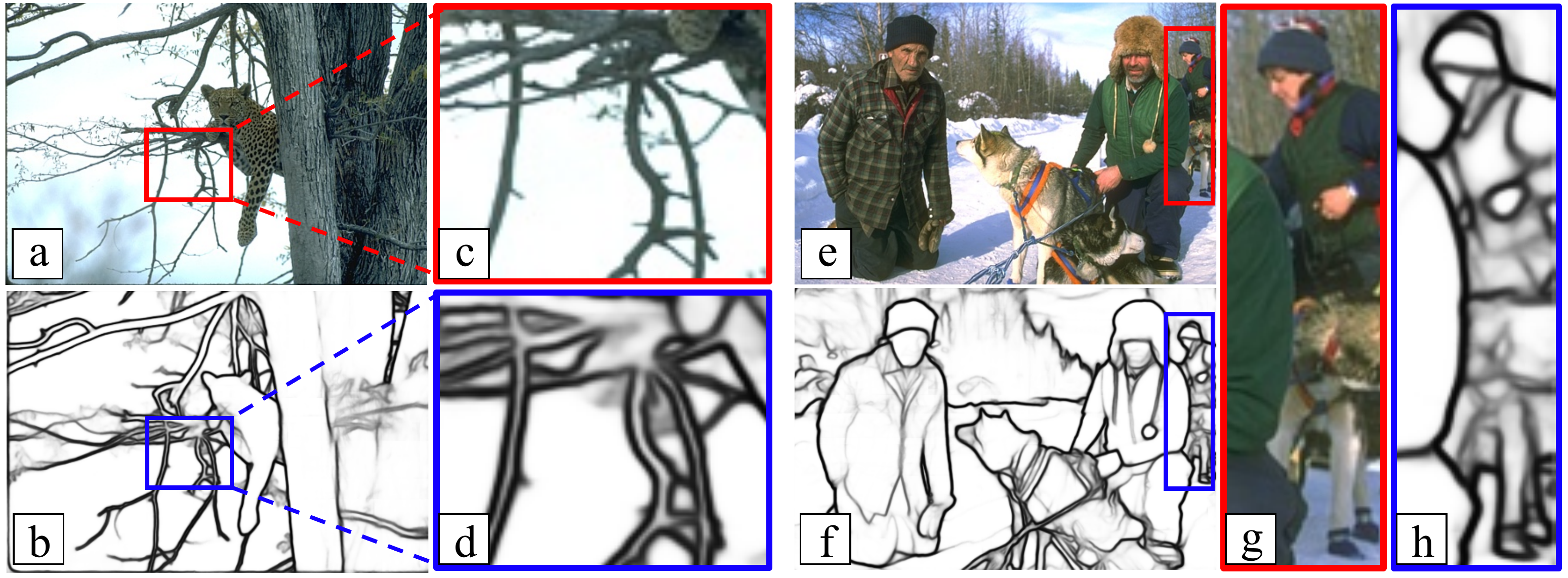}
\caption{\textbf{Examples of edge detection.} Our method, EDTER, extracts clear boundaries and edges by exploiting both global and local cues. (a, e): Input images from BSDS500~\cite{arbelaez2010bsds}. (b,f): Detected edges by EDTER. (c,d,g,h): Zoomed-in patches.}
\label{fig:figure1}
\vspace{-15pt}
\end{figure}

Inspired by the recent success of vision transformers~\cite{dosovitskiy2020image16x16,zheng2021rethinking,wang2021end,chen2021transformertracking}, especially their capability of modeling long-range contextual information, we propose to tailor transformers for edge detection. Two main challenges, however, need to be solved. Firstly, transformers are often applied to patches with a relatively large size due to computation concerns, while coarse-grained patches are unfavorable for learning accurate features for edges. It is crucial to perform self-attention on fine-grained patches without increasing the computational burden. Second, as shown in Fig.~\ref{fig:figure1} (d), extracting precise edges from intersected and thin objects is challenging. So it is necessary to design an effective decoder for generating edge-aware high-resolution features.

\begin{figure*}[!t]
\setlength{\abovecaptionskip}{3pt}
\setlength{\belowcaptionskip}{0pt}
\centering
\includegraphics[width=0.998\linewidth
]{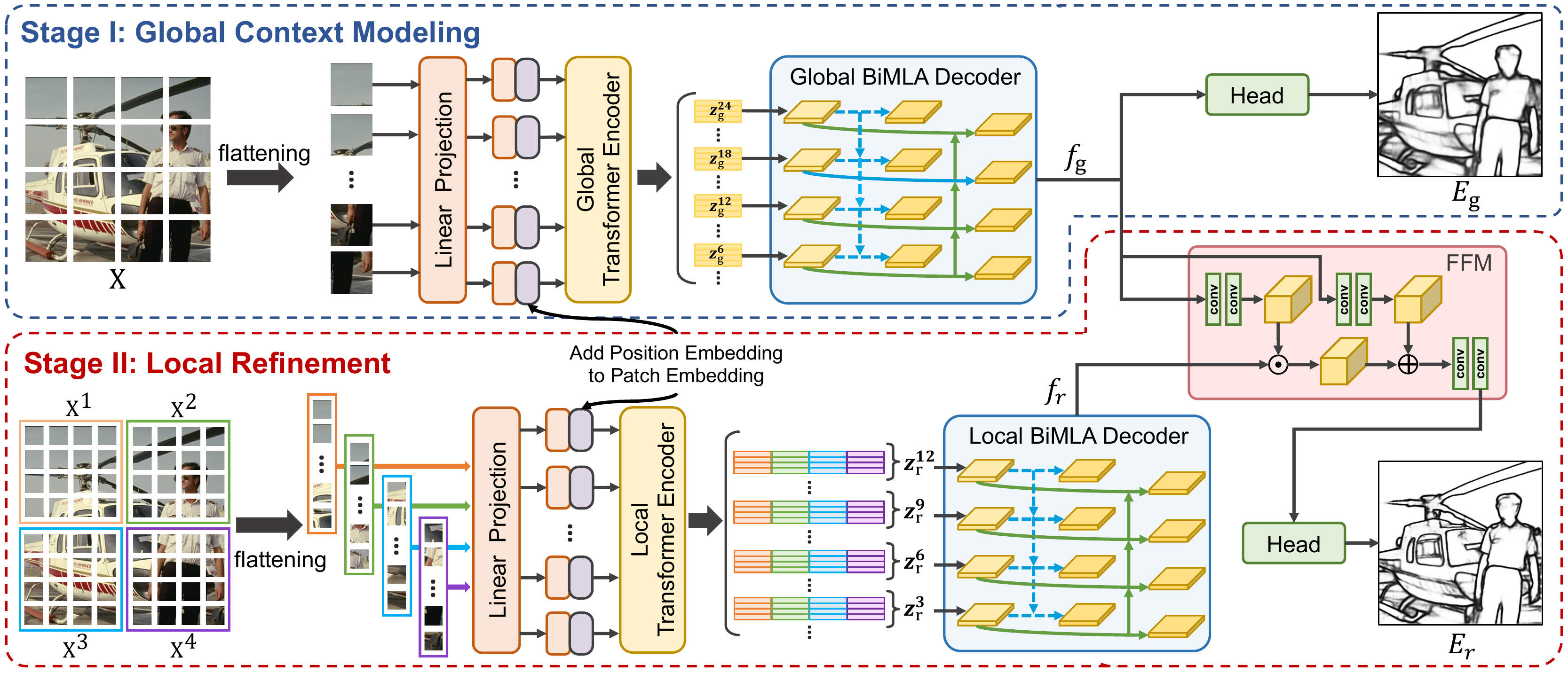}
\caption{\textbf{Overall framework.} In Stage I, we first feed the image into a global transformer encoder to compute the global attentions. Then, a global BiMLA decoder (see Fig.~\ref{fig:decoder}) generates the high-resolution features that are used to predict the edge maps via a decision head. In Stage II, similar to Stage I, the partitioned patches are inputted into a local transformer encoder to generate the local attentions. The concatenated attentions are utilized to decode the high-resolution features. At last, a decision head predicts the edge maps with the features of Stage I and Stage II fused by FFM.}
\label{fig:framework}
\vspace{-14pt}
\end{figure*}

To address the above issues, we develop a two-stage framework (Fig.~\ref{fig:framework}),  named Edge Detection TransformER (EDTER), to explore global context information and excavate fine-grained cues in local regions. In stage I, we split the image into coarse-grained patches and run a global transformer encoder on them to capture long-range global context. Then, we develop a novel Bi-directional Multi-Level Aggregation (BiMLA) decoder to generate high-resolution representations for edge detection. In stage II, we first divide the whole image into multiple sequences of fine-grained patches by sampling with a non-overlapping sliding window. Then a local transformer works on each sequence in turn to explore the short-range local cues. Afterward, all local cues are integrated and fed into a local BiMLA decoder to achieve the pixel-level feature maps. Finally, the information from both stages is fused by a Feature Fusion Module (FFM) and then is fed into a decision head to predict the final edge map. With the above efforts, EDTER can generate crisp and less noisy edge maps (Fig.~\ref{fig:figure1}). 

Our contributions are summarized as follows: (1) We propose a novel transformer-based edge detector, Edge Detection TransformER (EDTER), to detect object contours and meaningful edges in natural images. To our best knowledge, it is the first transformer-based edge detection model. (2) EDTER is designed to effectively explore long-range global context (Stage I) and capture fine-grained local cues (Stage II). Moreover, we propose a novel Bi-directional Multi-Level Aggregation (BiMLA) decoder to boost the information flow in the transformer. (3) To effectively integrate the global and local information, we use a Feature Fusion Module (FFM) to fuse the cues extracted from Stage I and Stage II. (4) Extensive experiments demonstrate the superiority of EDTER over the state-of-the-art methods on three well-known edge detection benchmarks, including BSDS500, NYUDv2, and Multicue.

%----------------------------------------------------------------
\section{Related Work}
As a fundamental task in computer vision, edge detection has been extensively studied over years. In the following, we highlight two lines of works most related to ours.

\vspace{0.2em}
\noindent{\bf Edge Detection.} Early edge detectors~\cite{kittler1983accuracy, canny1986computational, winnemoller2011xdog}, such as Sobel~\cite{kittler1983accuracy} and Canny~\cite{canny1986computational}, focus on analysing the image gradients to extract the edges. These methods provide elementary low-level cues and are widely used in computer vision applications. Learning-based methods~\cite{martin2004learning,dollar2006supervised, lim2013sketch} tend to integrate different low-level features and train a classifier to obtain boundaries and edges. Although these approaches achieve impressive performance compared to early works, they are based on hand-crafted features, limiting the ability to detect semantic boundaries and meaningful edges.

Recently, convolutional neural networks (CNNs) have been successfully introduced the edge detection study~\cite{kokkinos2015pushing, deng2020dscd, xu2017AMHNet, shen2015deepcontour, bertasius2015deepedge, bertasius2015hfl, maninis2016cob, deng2018lpcb, kelm2019rcn, poma2020dexined, PuHGL21iccv}. DeepEdge~\cite{bertasius2015deepedge} exploits object-aware cues extracted by multi-level CNN for contour detection. The method in~\cite{shen2015deepcontour} first partitions contour patches into sub-classes and then learns model parameters to fit each subclass. {More recently, some approaches improve edge detection~\cite{xie2015hed,liu2016rds, liu2017rcf, he2019bdcn, xu2017AMHNet}, segmentation~\cite{Tao2020hierarchical,chen2017deeplab,ZhaoPSPN17}, and object detection~\cite{LinFPN17} by using hierarchical multi-scale features.} Inspired by the seminal work of~\cite{xie2015hed}, most edge detectors~\cite{liu2016rds, liu2017rcf, he2019bdcn, xu2017AMHNet} generate object boundaries from hierarchical features by multi-level learning. Specifically, HED~\cite{xie2015hed} learns rich hierarchical features by performing supervisions on side output layers, which boosts the performance of edge detection. RCF~\cite{liu2017rcf} combines hierarchical features from all convolutional layers into a holistic architecture. To achieve effective results, BDCN~\cite{he2019bdcn} uses layer-specific supervision inferred from a bi-directional cascade structure to guide the training of each layer. PiDiNet~\cite{Su2021pidiNet} integrates the traditional edge detection operators into a CNN model for enhanced performance.

\vspace{0.2em}
\noindent{\bf Vision transformer.} First introduced to handle natural language tasks~\cite{devlin2019bert, lan2020albert, vaswani2017attention}, transformer is later extended to vision tasks owing to its capacity in modeling long-range dependencies including image classification~\cite{dosovitskiy2020image16x16}, semantic segmentation~\cite{zheng2021rethinking}, and object detection~\cite{carion2020detr}. Recently, it is applied in conjunction with CNN in DETR~\cite{carion2020detr} and the other variants~\cite{Li_2021_CVPR_Pose,Dai_2021_CVPR_UPDETR,Kim_2021_CVPR_HOTR,Wang_2021_CVPR_Tracker,Zou_2021_CVPR_HOI}. More recently, vision transformer (ViT)~\cite{dosovitskiy2020image16x16} directly uses the transformer to the sequences of image patches and achieves the state-of-the-art. This architecture brings direct inspiration to other computer vision tasks~\cite{zheng2021rethinking,liu2021swin,vaswani2021scaling,zhang2021multi,Li_2021_Diverse}. For example, SETR~\cite{zheng2021rethinking} shows superior accuracy in semantic segmentation using a pure transformer on image patches. These works demonstrate the effectiveness of transformers in capturing long-range dependencies and global context.

Our work is inspired by the above pioneer studies~\cite{dosovitskiy2020image16x16, zheng2021rethinking,liu2021swin}, but is significantly different in two aspects. First, the proposed EDTER, to the best of my knowledge, is the first usage of the transformer for generic edge detection. Second, our key idea is to learn the features that contain the global image context and fine-grained local cues by a two-stage framework with an affordable computational cost. With the integration of the global context and local cues, EDTER is superior in edge detection.

%----------------------------------------------------------------
\section{Edge Detection with Transformer}

\subsection{Overview}

The overall framework of the proposed EDTER is illustrated in Fig.~\ref{fig:framework}. EDTER explores the full image context information and fine-grained cues in two stages. In Stage I, we first split the input image into a sequence of coarse-grained patches and use a global transformer encoder to learn the global context information. Then a Bi-directional Multi-Level Aggregation (BiMLA) decoder is used to generate the high-resolution features. In Stage II, the whole image is divided into multiple sequences of fine-grained patches by sampling with a non-overlapping sliding window. Then we execute a local transformer encoder on each sequence in turn to capture short-range local cues. We integrate all local cues and input them into a local BiMLA decoder to achieve the pixel-level feature maps. Finally, the global and local features are integrated by a Feature Fusion Module (FFM) and then are fed into a decision head to predict the final edge maps.

\subsection{Review Vision Transformer}
The transformer encoders in our framework follow the vision transformer (ViT) in~\cite{dosovitskiy2020image16x16}, as briefly described below.

\vspace{0.4em}
\noindent{\bf Image Partition.} The first step in ViT is to transform a 2D image, denoted by $X \in \mathbb{R}^{H \times W \times 3}$, into a 1D sequence of image patches~\cite{dosovitskiy2020image16x16,zheng2021rethinking}. Concretely, we uniformly split $X$ into a sequence of flattened image patches of size $P \times P$, resulting in $\frac{H}{P} \times \frac{W}{P}$ vision tokens. Then, the sequence is mapped into a latent embedding space by a learnable linear projection. The projected features are called patch embeddings. Further, to preserve positional information, the standard learnable 1D position embeddings are added to the patch embeddings. Finally, the combined embeddings (denoted as $z^0$) are fed into the transformer encoder.

\begin{figure*}[!t]
\setlength{\abovecaptionskip}{3pt}
\setlength{\belowcaptionskip}{0pt}
\centering
\includegraphics[width=0.98\linewidth]{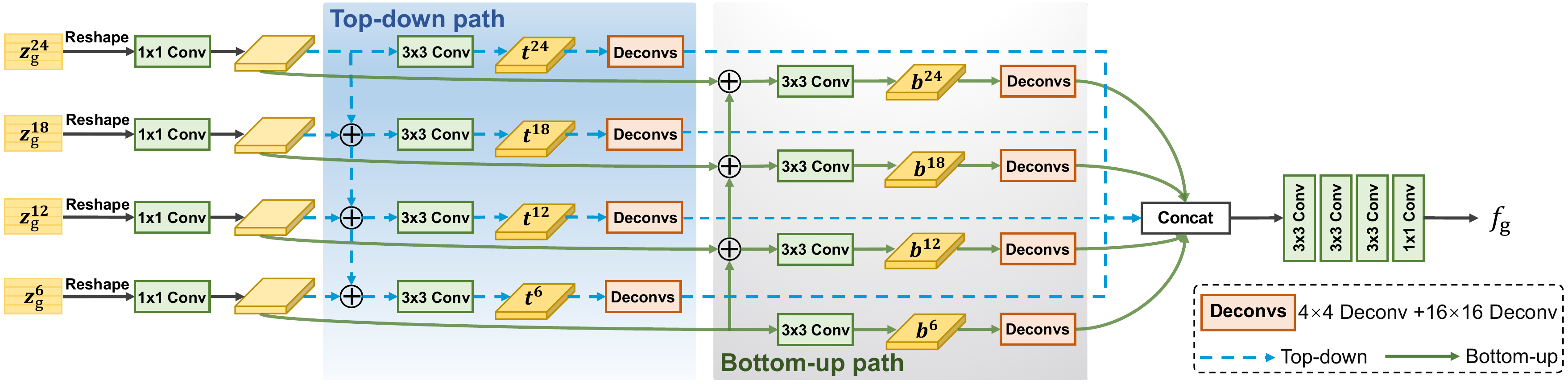}
\caption{The detailed architecture of the BiMLA decoder consists of a top-down path and a bottom-up path.}
\label{fig:decoder}
\vspace{-12pt}
\end{figure*}

\vspace{0.4em}
\noindent{\bf Transformer Encoder.} The standard transformer encoder~\cite{vaswani2017attention} consists of $L$ transformer blocks. Each block has a multi-head self-attention operation (MSA), a multi-layer perceptron (MLP), and two Layernorm steps (LN). Moreover, a residual connection layer is applied after each block. Generally, MSA performs $M$ self-attentions in parallel and projects their concatenated outputs. In the $m^{th}$ self-attention, given the output $z^{l-1}\in \mathbb{R}^{N \times C}$ of the $(l-1)^{th}$ transformer block, the queries $Q \in \mathbb{R}^{N \times U}$, keys $K \in \mathbb{R}^{N \times U}$, and values $V \in \mathbb{R}^{N \times U}$ are computed by
\begin{equation}
    Q = \hat{z}^{l-1}W_Q,\quad K = \hat{z}^{l-1}W_K,\quad V = \hat{z}^{l-1}W_V,
\end{equation}
where $\hat{z}^{l-1}={\rm LN}(z^{l-1})$, $W_Q, W_K,W_V \in \mathbb{R}^{C \times U}$ are the parameter matrices, $C$ is the dimension of embeddings, and $U$ is the dimension of $Q$, $K$, and $V$. Then, we compute the output of the $m^{th}$ self-attention based on the pairwise similarity between two elements of the sequence by
\begin{equation}
    y^m_{\rm sa}={\rm softmax} \Big(\frac{Q{K}^\mathsf{T}}{\sqrt{d}}\Big)V.
\end{equation}
where $y^m_{\rm sa}$ is the computed attention weight. Finally, MSA can be formulated as
\begin{equation}
    y_{\rm msa}={\rm MSA}(\hat{z}^{l-1})=\big[y^1_{\rm sa},y^2_{\rm sa},\dots, y^M_{\rm sa}\big] \ W_{\rm O},
\end{equation}
where $y_{\rm msa}$ is the output of MSA, $W_{\rm O}\in \mathbb{R}^{M \cdot U \times C}$ represents the projection parameters, and $[\cdot]$ is the concatenation. In this work, we fix $M = 16$ following the setting in~\cite{dosovitskiy2020image16x16}.

\subsection{Stage I: Global Context Modeling} 
Generally, edges and boundaries in images are defined to be semantically meaningful. It is crucial to capture the abstract cues and the global context of the whole image. In the first stage, we explore the global contextual features on coarse-grained patches by a global transformer encoder $\mathcal{G}_{\rm E}$ and a global decoder $\mathcal{G}_{\rm D}$.

Specifically, we first split the input image into a sequence of coarse-grained patches of size 16$\times$16, and then generate the embeddings $z^0_{\rm g}$ that serve as input of the encoder. Next, the global transformer encoder $\mathcal{G}_{\rm E}$ works on the embeddings $z^0_{\rm g}$ to compute the global attentions,
\begin{equation}
    z_{\rm g} =\{z_{\rm g}^{1}, z_{\rm g}^{2}, \dots, z_{\rm g}^{L_{\rm g}}\} = \mathcal{G}_{\rm E}(z^0_{\rm g}),
\end{equation}
where $z_{\rm g}^{1}, z_{\rm g}^{2}, \dots, z_{\rm g}^{L_{\rm g}} \in \mathbb{R}^{\frac{HW}{256} \times C}$ represent the outputs of successive blocks in $\mathcal{G}_{\rm E}$, and $L_{\rm g}$ is the number of transformer blocks in $\mathcal{G}_{\rm E}$. In our experiments, we set $\mathcal{G}_{\rm E}$ to 24 following~\cite{dosovitskiy2020image16x16}. Next, the sequence of global context features $z_{\rm g}$ are upsampled to high-resolution features by the global decoder $\mathcal{G}_{\rm D}$ for incorporation.

\vspace{0.3em}\noindent{\bf BiMLA Decoder.} 
It is crucial to generate edge-aware pixel-level representations for detecting precise and thin edges. Thus, we expect to design a practical decoder that can encourage the transformer encoder to compute the edge-aware attentions and upsample the attentions in a learnable manner. Inspired by the multi-level feature aggregation in vision tasks~\cite{LinFPN17,xie2015hed,liu2017rcf,liu2016rds,he2019bdcn,zheng2021rethinking}, we propose a novel Bi-directional Multi-Level Aggregation (\textbf{BiMLA}) decoder, as illustrated in Fig.~\ref{fig:decoder}, to achieve the goal.

In BiMLA, a bi-directional feature aggregation strategy is designed that includes a top-down path and a bottom-up path to boost the information flow in the transformer encoder. More specifically, we first uniformly divide $L_{\rm g}$ transformer blocks into four groups, and take the embedding features $\{z_{\rm g}^{6}, z_{\rm g}^{12}, z_{\rm g}^{18}, z_{\rm g}^{24}\}$ from the last block of each group as inputs. Then we reshape them to 3D features with the size of $\frac{H}{16}\times\frac{W}{16}\times C$. For the top-down path, we attach the same design (one 1$\times$1 convolutional layer and one 3$\times$3 convolutional layer) to each reshaped feature and obtain four output features $t^{6},t^{12},t^{18},t^{24}$, following the way of SETR-MLA~\cite{zheng2021rethinking}. Likewise, the bottom-up path starts from the lowest level (\ie, $z_{\rm g}^{6}$) and gradually approaches the top level (\ie, $z_{\rm g}^{24}$) by attaching one 3$\times$3 convolutional layer on multi-level features, and finally produce another four output features $b^{6},b^{12},b^{18},b^{24}$. Besides, unlike SETR-MLA~\cite{zheng2021rethinking} that upsamples the features via bilinear operation, our BiMLA passes each aggregated feature through a deconvolutional block, contains two deconvolutional layers with 4$\times$4 kernels and 16$\times$16 kernels, respectively. Each deconvolutional layer is followed by Batch Normalization (BN) and ReLU operations. The eight upsampled features from the bi-directional path are then concatenated into one tensor. Moreover, BiMLA uses an additional stack of convolutional layers to smooth the concatenated features. The stack consists of three 3$\times$3 convolutional layers and one 1$\times$1 convolutional layer with BN and ReLU. The process of BiMLA decoder is formulated as 
\begin{equation}
    f_{\rm g} = \mathcal{G}_{\rm D}(z_{\rm g}^{6}, z_{\rm g}^{12}, z_{\rm g}^{18}, z_{\rm g}^{24}),
\end{equation}
where $f_{\rm g}$ is the pixel-level global features, and $\mathcal{G}_{\rm D}$ represents the global BiMLA decoder. After obtaining the coarse-grained global context features, we will capture the fine-grained local context features in the next stage.

\subsection{Stage II: Local Refinement} 
It is essential to explore fine-grained context features for pixel-level predictions, especially for edge detection. The ideal edge width is one pixel, while 16$\times$16 patches are not conducive to extracting thin edges. Taking pixels as tokens sounds an intuitive remedy, however, it is practically infeasible due to heavy computational cost. Our solution is to use a non-overlapping sliding window to sample the image and then calculate the attentions within the sampled regions. The number of patches in the window is fixed, so the computational complexity is linearly related to the image size.

Thus motivated, we propose to capture the short-range fine-grained context features in Stage II, as shown at the bottom of Fig.~\ref{fig:framework}. In particular, we perform the non-overlapping sliding window with a size of $\frac{H}{2} \times \frac{W}{2}$ on image $X \in \mathbb{R}^{H \times W \times 3}$, and the input image $X$ is decomposed into a sequence $\{ X^{1}, X^{2}, X^{3}, X^{4} \}$. For each window, we split it into fine-grained patches of size 8$\times$8 and compute the attentions by a shared local transformer encoder $\mathcal{R}_{\rm E}$. Then we concatenate the attentions of all windows to obtain $z_{\rm r}=\{z_{\rm r}^1,\dots,z_{\rm r}^{L_{\rm r}}\} \in \mathbb{R}^{\frac{HW}{64}\times C}$. To further economize the computing resource, we set $L_{\rm r} =12$ that means the local transformer encoder consists of $12$ transformer blocks. Similar to global BiMLA, we evenly select $\{z_{\rm r}^3, z_{\rm r}^6, z_{\rm r}^9, z_{\rm r}^{12}\}$ from $z_{\rm r}$ and input them into the local BiMLA ${\mathcal{R}_D}$ to generate the local features with high-resolution,
\begin{equation}
    f_{\rm r} = \mathcal{R}_{\rm D}(z_{\rm r}^3, z_{\rm r}^6, z_{\rm r}^9, z_{\rm r}^{12}),
\end{equation}
where $f_{\rm r}$ indicates the local features. Different from global BiMLA, we replace the 3$\times$3 convolutional layer with the 1$\times$1 convolutional layer in local BiMLA, so as to avoid artificial edges caused by the padding operation.

\vspace{0.5em}
\noindent{\bf Feature Fusion Module.} Finally, we incorporate the context cues from both levels by a Feature Fusion Module (FFM) and predict the edge maps by a local decision head. FFM takes the global context as the prior knowledge and modulates the local context, which produces the fusion features containing global context and fine-grained local details. As shown in Fig.~\ref{fig:framework}, FFM consists of a spatial feature transform block~\cite{wang2018sft} and two 3$\times$3 convolutional layers followed by BN and ReLU operations. The former is for modulating, and the latter is for smoothing. Then the fusion features are fed into the local decision head $\mathcal{R}_{\rm H}$ to predict the edge maps $E_{\rm r}$, 
\begin{equation}
    E_{\rm r}=\mathcal{R}_{\rm H} \big ({\rm FFM}(f_{\rm g}, f_{\rm r}) \big),
\end{equation}
where $\mathcal{R}_{\rm H}$ is the local decision head that consists of a 1$\times$1 convolutional layer and a sigmoid operation.

\subsection{Network Training} 
\label{training}
To train the two-stage framework EDTER, we first optimize Stage I to generate global features that represent the whole image context information. Then, we fix the parameters of Stage I and train Stage II to generate edge maps. 

\vspace{0.5em}
\noindent{\bf Loss Function.} We employ the loss function proposed in~\cite{xie2015hed} for each edge map. Given an edge map $E$ and the corresponding ground truth $Y$, the loss is calculated as
\begin{equation}
\begin{split}\ell \left (E,Y \right) & = - \sum_{i,j} \big( Y_{i,j}\alpha{\rm log}(E_{i,j}) \\
    &  + (1-Y_{i,j})(1-\alpha){\rm log}(1-E_{i,j}) \big), 
\end{split}
\end{equation}
where $E_{i,j}$ and $Y_{i,j}$ are the $(i,j)^{th}$ element of matrix $E$ and $Y$, respectively. Moreover, $\alpha=|Y^-|/(|Y^-|+|Y^+|)$ indicates the percentage of negative pixel samples, where $|\cdot|$ denotes the number of pixels. In practice, the annotations of BSDS500~\cite{arbelaez2010bsds} are labeled by multiple annotators. Inconsistent annotations lead to problematic convergence behavior~\cite{xie2015hed}. Following~\cite{liu2017rcf}, we first normalize multiple labels to an edge probability map with ranges $[0, 1]$, and then use a threshold $\eta$ to select pixels. The pixel is marked as a positive sample if the probability value is higher than $\eta$; otherwise, it is indicated as a negative sample.

\vspace{0.5em}
\noindent{\bf Training Stage I.} For training Stage I, we first incorporate the global decision head on the global feature maps to generate the edge maps $E_{\rm g}$ by
\begin{equation}
    E_{\rm g}=\mathcal{G}_{\rm H}(f_{\rm g}),
\end{equation}
where $\mathcal{G}_{\rm H}$ indicates the global decision head that consists of a 1$\times$1 convolutional layer and a sigmoid layer. Moreover, we obtain multiple side outputs $S^{1}_{\rm g}, S^{2}_{\rm g}, \dots ,S^{8}_{\rm g}$ by performing the same design (a 4$\times$4 deconvolutional layer and a 16$\times$16 deconvolutional layer) to the intermediate features $t^{6},t^{12},t^{18},t^{24}$ and $b^{6},b^{12},b^{18},b^{24}$ extracted by the global BiMLA decoder, which progressively enforce the encoder to emphasize edge-aware attentions.

Stage I is optimized by minimizing the losses between each edge map and the ground truth. The loss function of Stage I is formulated as
\begin{equation}
\mathcal{L}_{\rm g} = \mathcal{L}_{\rm g}^{E} + \lambda\mathcal{L}_{\rm g}^{\rm side}
=\ell \left (E_{\rm g},Y \right)+\lambda\sum_{k=1}^{8} \ell \left(S_{\rm g}^k, Y \right),
\end{equation}
where $\mathcal{L}_{\rm g}^{E}$ is the loss for $E_{\rm g}$, $\mathcal{L}_{\rm g}^{\rm side}$ denotes side loss, and $\lambda$ is the weight for balancing $\mathcal{L}_{\rm g}^{E}$ and $\mathcal{L}_{\rm g}^{\rm side}$. In our experiments, we set $\lambda$ to 0.4.

\vspace{0.5em}
\noindent{\bf Training Stage II.} After training Stage I, we fix the parameters of Stage I and move on to Stage II. Similar to the training of Stage I, we perform the same operation (a 4$\times$4 deconvolutional layer and an 8$\times$8 deconvolutional layer) on the intermediate features extracted from the local BiMLA decoder to generate the side outputs $S^{1}_{\rm r}, S^{2}_{\rm r}, \dots ,S^{8}_{\rm r}$. Finally, the loss function of Stage II is defined as
\begin{equation}
\mathcal{L}_{\rm r} = \mathcal{L}_{\rm r}^{E} + \lambda\mathcal{L}_{\rm r}^{\rm side}
=\ell \left (E_{\rm r},Y \right)+\lambda\sum_{k=1}^{8} \ell \left(S_{\rm r}^k, Y \right),
\end{equation}
where $\mathcal{L}_{\rm r}^{E}$ and $\mathcal{L}_{\rm r}^{\rm side}$ are the losses for $E_{\rm r}$ and side outputs, respectively. We again set $\lambda=0.4$.

%-----------------------------------------------------------------
\section{Experiments}

\subsection{Datasets}
We conduct the experiments on three popular benchmarks: BSDS500~\cite{arbelaez2010bsds}, NYUDv2~\cite{silberman2012indoor} and Multicue~\cite{mely2016multicue}. 

\textbf{BSDS500}~\cite{arbelaez2010bsds} contains 500 RGB natural images, 200 for training, 100 for validation, and 200 for testing. Each image is manually annotated by five different subjects on average. Our model is trained on the training and validation sets and evaluated on the testing set. Similar to~\cite{xie2015hed,liu2017rcf,he2019bdcn}, we augment the dataset by rotating each image at 16 different angles and flipping the image at each angle. Moreover, most previous works~\cite{liu2017rcf,liu2016rds,wang2017ced,he2019bdcn} use PASCAL VOC Context Dataset~\cite{everingham2010pascal} as the additional training data, which provides full-scene segmentation annotations with more than 400 classes, and consists of 10,103 images for training. The outside boundaries extracted from the segmentation annotations are beneficial to infer semantic and context cues in Stage I. Therefore, we first pre-train Stage I on PASCAL VOC Context Dataset~\cite{everingham2010pascal} and then fine-tune it on BSDS500~\cite{arbelaez2010bsds}. The PASCAL VOC Context Dataset~\cite{everingham2010pascal} is only used for training Stage I.

\textbf{NYUDv2}~\cite{silberman2012indoor} contains 1,449 pairs of aligned RGB and depth images, and it is split into 381 training, 414 validation, and 654 testing images. Following~\cite{xie2015hed,liu2017rcf}, we combine the training and validation sets as the training data, and then augment them by rotating the images and annotations to 4 different angles, randomly flipping, and scaling.
 
\textbf{Multicue}~\cite{mely2016multicue} is composed of 100 challenging natural scenes captured by a binocular stereo camera. Each scene contains a left-view and a right-view short sequences. The last frame of left-view sequences from each scene is labeled with edges and boundaries. Following~\cite{xie2015hed,liu2017rcf,he2019bdcn}, we randomly select 80 images for training and the remaining 20 images for testing. We repeat the process three times and average the scores of three independent trials as the final results. The data augmentation follows~\cite{xie2015hed,liu2017rcf}.

\begin{figure}[t]
\setlength{\abovecaptionskip}{3pt}
\setlength{\belowcaptionskip}{0pt}
\centering
\includegraphics[width=0.95\linewidth,height=.35\linewidth]{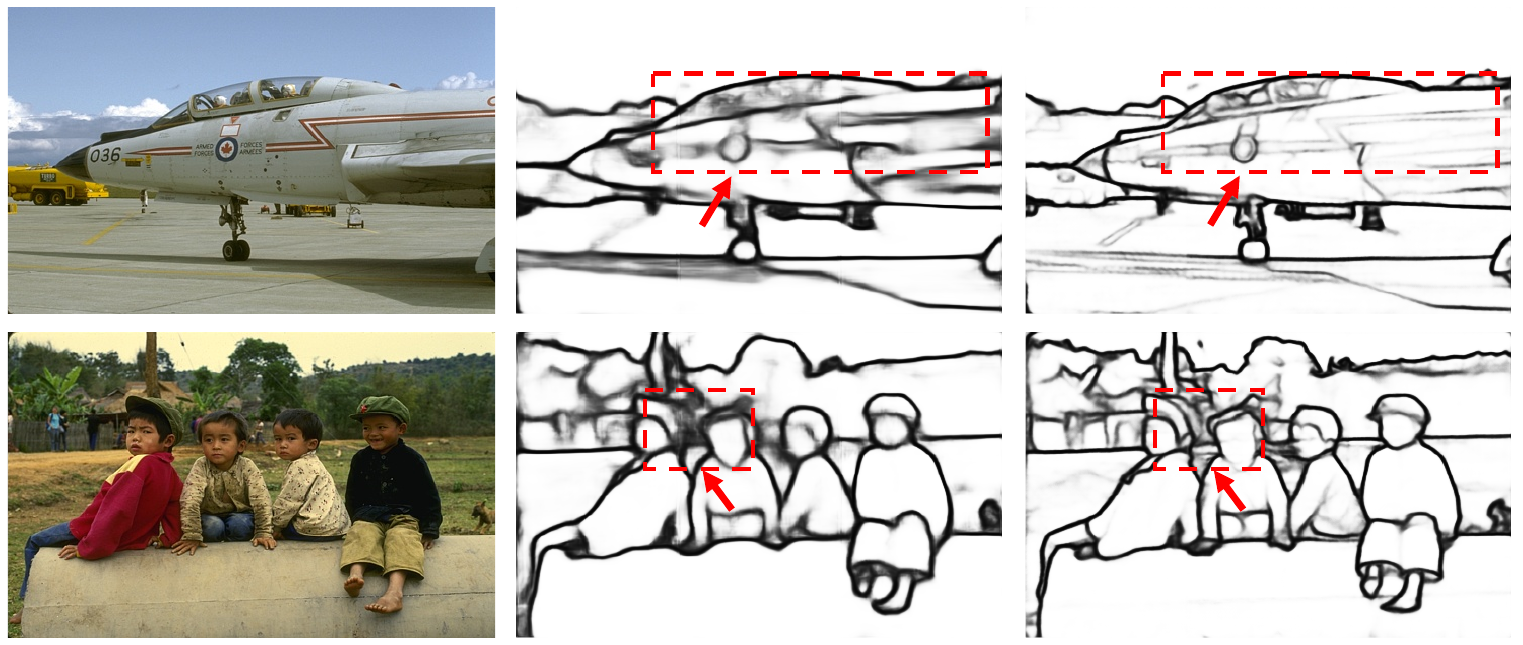}
\caption{Qualitative comparison of different decoders in EDTER on BSDS500. From left to right are the input images, the results of EDTER using SETR-MLA and BiMLA decoder, respectively.}
\label{fig:ablation_fig}
\vspace{-14pt}
\end{figure}

\subsection{Implementation Details}
We implement our EDTER using PyTorch~\cite{paszke2017automatic}. We initialize the transformer blocks of our model using the pre-trained weights by ViT~\cite{dosovitskiy2020image16x16}. We set the threshold $\eta$ as 0.3 to select positive samples for BSDS500 and Multicue Edge, and 0.4 for Multicue Boundary. Each image has only one annotation in NYUDv2, thus no $\eta$ is needed. We use SGD optimizer with momentum=0.9 and weight decay=2e-4, and adopt a polynomial learning rate decay schedule~\cite{Zhao17PyramidScene} on all datasets. The initial learning rate is set as 1e-6 for BSDS500, NYUDv2 and Multicue Boundary, and 1e-7 for Multicue Edge. During training, we set the same iteration numbers for both stages. Specially, we train 80k iterations for BSDS500 and Multicue boundary, 40k for NYUDv2, and 4k for Multicue Edge. Each image is randomly cropped to 320$\times$320 in training. Compared with BSDS500, the annotations of NYUDv2 are unitary, and the scale of Multicue is small, which quickly overfits of the model trained on them. Therefore, we set the batch size to 8 for BSDS500, and 4 for NYUDv2 and Multicue. All the experiments are conducted on a V100 GPU. The training of EDTER takes about 26.4 hours (15.1 for Stage I and 11.3 for Stage II). The inference runs at 2.2 \textit{fps} on a V100. During Training, the GPU consumption of Stage I and Stage II are about 15GB and 14GB for 320$\times$320 images respectively. Besides, EDTER brings 332.0G FLOPs in Stage I and 470.25G FLOPs in Stage II.

During evaluation, we record three metrics for all datasets: fixed contour threshold (ODS), per-image best threshold (OIS), and average precision (AP). Moreover, a non-maximum suppression~\cite{canny1986computational} is performed on the predicted edge maps before evaluation. Following previous works~\cite{xie2015hed,liu2017rcf}, the localization tolerance controls the maximum allowed distance in matches between edge results and the ground truth, which is set to 0.0075 for BSDS500 and Multicue, and 0.011 for NYUDv2.

\begin{table}%[!h]
\setlength{\abovecaptionskip}{0pt}
\caption{Ablation study of the effectiveness of the proposed two-stage strategy, BiMLA decoder and FFM in EDTER on BSDS500. All results are computed with a single scale input.}
\centering
\footnotesize
\renewcommand\arraystretch{0.9}
\renewcommand\tabcolsep{5.8pt}
\begin{tabular}{c|cc|c|ccc}
    \toprule
    \multirow{2}{*}{Stage}   & \multicolumn{2}{c|}{Decoder}  & \multirow{2}{*}{FFM}   & \multirow{2}{*}{ODS}  & \multirow{2}{*}{OIS}  & \multirow{2}{*}{AP} \\
    \cline{2-3}
    &SETR-MLA        &BiMLA      & & & & \\
    \toprule
    \multirow{2}{*}{I}
    &$\surd$    &$\times$   
    &\multirow{2}{*}{\diagbox{\textcolor{white}{.}}{\textcolor{white}{.}}}    & 0.790     & 0.806     & 0.836\\
    &$\times$   &$\surd$    &    & {\bf 0.817}   & {\bf 0.835}   & {\bf 0.867}\\
    \toprule
    \multirow{3}{*}{II}
    &$\surd$    &$\times$   &$\surd$    & 0.799     & 0.816     & 0.848\\
    &$\times$   &$\surd$    &$\surd$    & {\bf 0.824}  & {\bf 0.841}  & {\bf 0.880}\\
    &$\times$   &$\surd$    &$\times$   & 0.820     & 0.835     & 0.867\\
    \bottomrule
\end{tabular}
\vspace{-12pt}
\label{tab:decoder_ffm}
\end{table}

\subsection{Ablation Study}

\noindent{\bf Effectiveness of key components in EDTER.} We first conduct experiments to verify the impact of key EDTER components: BiMLA and FFM. The quantitative results are summarized in Table~\ref{tab:decoder_ffm}. First, the performance of ODS, OIS, AP is largely improved (about 2.5\%, 3\%, 3\%) by BiMLA compared with SETR-MLA~\cite{zheng2021rethinking} in both stages. The performance of Stage II significantly surpasses Stage I under either decoder. It illustrates that the two-stage strategy fuses more critical information for edge detection. Besides, we present the predicted edge maps by the SETR-MLA and BiMLA decoders shown in Fig.~\ref{fig:ablation_fig}. With the BiMLA decoder, EDTER can accurately detect edges in some local areas (red bounding boxes) and produce less noisy edges. To verify the effectiveness of FFM, we remove the FFM and directly concatenate the feature maps from two stages to construct a variant of EDTER. Without using FFM (row 5), the scores drop by 0.4\%, 0.6\%, and 1.3\% in ODS, OIS, and AP, respectively.

\begin{figure}[t]
\setlength{\abovecaptionskip}{3pt}
\setlength{\belowcaptionskip}{0pt}
\centering
\includegraphics[width=0.95\linewidth,height=.35\linewidth]{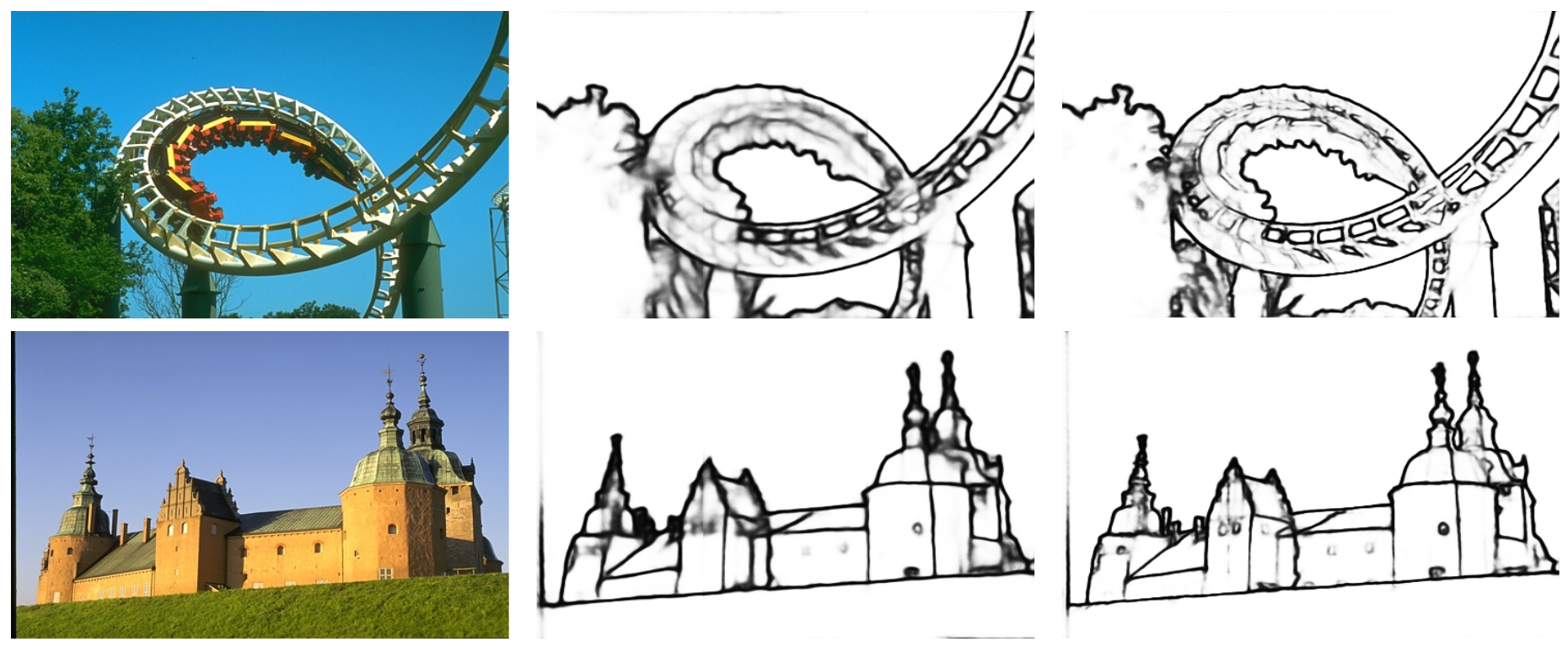}
\caption{Qualitative comparison of different stages in EDTER on BSDS500. From left to right are the input images, the results of EDTER-Stage I and EDTER-Stage II, respectively.}
\label{fig:ablation_2stage}
\vspace{-7pt}
\end{figure}

\noindent{\bf Effectiveness of stages and patch size.} We run ablation experiments to verify the effectiveness of the two-stage strategy. The comparative results are presented in Table~\ref{tab:stage_size}. Compared with Stage I (row 1), we add the second stage and set the patch size to 8$\times$8 (row 2), which obtains the performance gain by 0.7\%, 0.6\%, 1.3\% in ODS, IS, and AP, respectively. Moreover, as visualized in Fig.~\ref{fig:ablation_2stage}, the predicted edges of Stage II are more clear and crisp in some local details. It shows that the effectiveness of the two-stage strategy for edge detection. Then, to analyze the impact of the patch size, we create a variant that uses the patch size of 4$\times$4 in Stage II (row 3). Concretely, in Stage II, the image is first decomposed into a sequence $\{X^1,X^2,\dots,X^{16}\}$ by the sliding window with a size of $\frac{H}{4}\times\frac{W}{4}$, and then we split each window into a sequence of 4$\times$4 patches and generate local cues. Compared with row 2 using 8$\times$8 patches in Stage II, the performance is slightly improved. Moreover, we report experiments with EDTER model variants that use more stages to capture local context, obtained by adding Stage III and setting the patch size to 4$\times$4. In Table~\ref{tab:stage_size} (row 4), the scores are marginally increased by fusing context cues of three stages. Since the edge extracted from the networks inevitably occupies multiple pixels, 4$\times$4 patches seldom bring significant gains. Considering the trade-off between computational efficiency and performance, we employ the setting of \textit{16$\times$16 in Stage I} and \textit{8$\times$8 in Stage II} to perform all subsequent experiments.

\begin{table}%[t!]
\setlength{\abovecaptionskip}{0pt}
\caption{Quantitative results of EDTER with training on different patch sizes and stage numbers on BSDS500. ``-'' means Stage III is not extended. All results are computed with a single scale input.}
\centering
\footnotesize
\renewcommand\arraystretch{0.9}
\renewcommand\tabcolsep{8.2pt}
\begin{tabular}{c|c|c|ccc}
    \toprule
    \multicolumn{3}{c|}{Patch Size}  & \multirow{2}{*}{ODS}  & \multirow{2}{*}{OIS}  & \multirow{2}{*}{AP} \\
    \cline{1-3}
    Stage I         &Stage II   &Stage III  &       &       & \\
    \toprule
    16$\times$16    &-          &-          &0.817  &0.835  &0.867\\
    16$\times$16    &8$\times$8 &-          &0.824  &0.841  &0.880\\
    16$\times$16    &4$\times$4 &-          &0.825  &0.843  &0.882\\
    16$\times$16    &8$\times$8 &4$\times$4 &0.826  &0.843  &0.883\\
    \bottomrule
\end{tabular}
\label{tab:stage_size}
\vspace{-14pt}
\end{table}

\subsection{Comparison with State-of-the-arts}

\begin{table}[tbp]
\setlength{\abovecaptionskip}{0pt}
\caption{Results on BSDS500~\cite{arbelaez2010bsds} testing set. The best two results are highlighted in \BEST{red} and \SBEST{blue}, respectively, and same for other tables. MS is the multi-scale testing, and VOC means training with extra PASCAL VOC data.}
\centering
\footnotesize
\renewcommand\arraystretch{0.9}
\renewcommand\tabcolsep{5.0pt}
\begin{tabular}{c|l|c|ccc}
    \toprule
    \multicolumn{2}{c|}{Method}          &\makecell[c]{Pub.'Year}     & ODS       & OIS       & AP    \\
    \toprule
    \multirow{9}{*}{\rotatebox{90}{Traditional Method}}
    & Canny~\cite{canny1986computational}     & PAMI'86     & 0.600     & 0.640     & 0.580 \\
    & Felz-Hutt~\cite{felzenszwalb2004felz}   & IJCV'04      & 0.610     & 0.640     & 0.560 \\
    & gPb-owt-ucm~\cite{arbelaez2010contour}  & PAMI'10     & 0.726     & 0.757     & 0.696 \\
    & SCG~\cite{xiaofeng2012scg}              & NeurIPS'12   & 0.739     & 0.758     & 0.773 \\
    & Sketch Tokens~\cite{lim2013sketch}      & CVPR'13      & 0.727     & 0.746     & 0.780 \\
    & PMI~\cite{isola2014pmi}                 & ECCV'14      & 0.741     & 0.769     & 0.799 \\
    & SE~\cite{dollar2014fast}                & PAMI'14     & 0.746     & 0.767     & 0.803 \\
    & OEF~\cite{hallman2015oef}               & CVPR'15      & 0.746     & 0.770     & 0.820 \\
    & MES~\cite{sironi2015mes}                & ICCV'15      & 0.756     & 0.776     & 0.756 \\
    \toprule
    \multirow{19}{*}{\rotatebox{90}{CNN-based Method}}
    & DeepEdge~\cite{bertasius2015deepedge}   & CVPR'15      & 0.753     & 0.772     & 0.807 \\
    & CSCNN~\cite{hwang2015cscnn}             & ArXiv'15       & 0.756     & 0.775     & 0.798 \\
    & MSC~\cite{sironi2015msc}                & PAMI'15     & 0.756     & 0.776     & 0.787 \\
    & DeepContour~\cite{shen2015deepcontour}  & CVPR'15      & 0.757     & 0.776     & 0.800 \\
    & HFL~\cite{bertasius2015hfl}             & ICCV'15      & 0.767     & 0.788     & 0.795 \\
    & HED~\cite{xie2015hed}                   & ICCV'15      & 0.788     & 0.808     & 0.840 \\
    & Deep Boundary~\cite{kokkinos2015pushing}& ICLR'15      & 0.813     & 0.831     & 0.866 \\
    & CEDN~\cite{yang2016cedn}                & CVPR'16      & 0.788     & 0.804     & - \\
    & RDS~\cite{liu2016rds}                   & CVPR'16      & 0.792     & 0.810     & 0.818 \\
    & COB~\cite{maninis2016cob}               & ECCV'16      & 0.793     & 0.820     & 0.859 \\
    & DCD~\cite{liao2017dcd}                  & ICME'17      & 0.799     & 0.817     & 0.849 \\
    & AMH-Net~\cite{xu2017AMHNet}             & NeurIPS'17   & 0.798     & 0.829     & 0.869 \\
    & RCF~\cite{liu2017rcf}                   & CVPR'17      & 0.811     & 0.830     & - \\
    & CED~\cite{wang2017ced}                  & CVPR'17      & 0.815     & 0.833     & 0.889 \\
    & LPCB~\cite{deng2018lpcb}                & ECCV'18      & 0.815     & 0.834     & - \\
    & BDCN~\cite{he2019bdcn}                  & CVPR'19      & 0.828     & 0.844     & 0.890 \\
    & DexiNed~\cite{poma2020dexined}          & WACV'20      & 0.729     & 0.745     & 0.583 \\
    & DSCD~\cite{deng2020dscd}                & ACMMM'20     & 0.822     &\SBEST{0.859}     & - \\
    & PiDiNet~\cite{Su2021pidiNet}            & ICCV'21      & 0.807     & 0.823     & - \\
    \toprule
    \multirow{4}{*}{\rotatebox{90}{Ours}}
    & EDTER                  & -      & 0.824           & 0.841         & 0.880 \\
    & EDTER-VOC              & -      & 0.832           & 0.847         & 0.886 \\
    & EDTER-MS               & -      & \SBEST{0.840}   & 0.858 & \SBEST{0.896} \\  
    & EDTER-MS-VOC           & -      & \BEST{0.848}    & \BEST{0.865}  & \BEST{0.903} \\
    \bottomrule
\end{tabular}
\label{tab:bsds}
\vspace{-9pt}
\end{table}

\vspace{0.5em}
\noindent{\bf On BSDS500.} We compare our model with \textit{traditional detectors} including Canny~\cite{canny1986computational}, Felz-Hutt~\cite{felzenszwalb2004felz}, gPb-owt-ucm~\cite{arbelaez2010contour}, SCG~\cite{xiaofeng2012scg}, Sketch Tokens~\cite{lim2013sketch}, PMI~\cite{isola2014pmi}, SE~\cite{dollar2014fast}, OEF~\cite{hallman2015oef} and MES~\cite{sironi2015mes}, and \textit{deep-learning-based detectors} including DeepEdge~\cite{bertasius2015deepedge}, CSCNN~\cite{hwang2015cscnn}, DeepContour~\cite{shen2015deepcontour}, HFL~\cite{bertasius2015hfl}, HED~\cite{xie2015hed}, Deep Boundary~\cite{kokkinos2015pushing}, CEDN~\cite{yang2016cedn}, RDS~\cite{liu2016rds}, COB~\cite{maninis2016cob}, DCD~\cite{liao2017dcd}, AMH-Net~\cite{xu2017AMHNet}, RCF~\cite{liu2017rcf}, CED~\cite{wang2017ced}, LPCB~\cite{deng2018lpcb}, BDCN~\cite{he2019bdcn}, DexiNed~\cite{poma2020dexined}, DSCD~\cite{deng2020dscd} and PiDiNet~\cite{Su2021pidiNet}. The best results of all the methods are taken from their publications.

\begin{figure}[tbp]
\setlength{\abovecaptionskip}{0pt}
\setlength{\belowcaptionskip}{0pt}
\centering
\includegraphics[width=0.85\linewidth,height=.71\linewidth]
{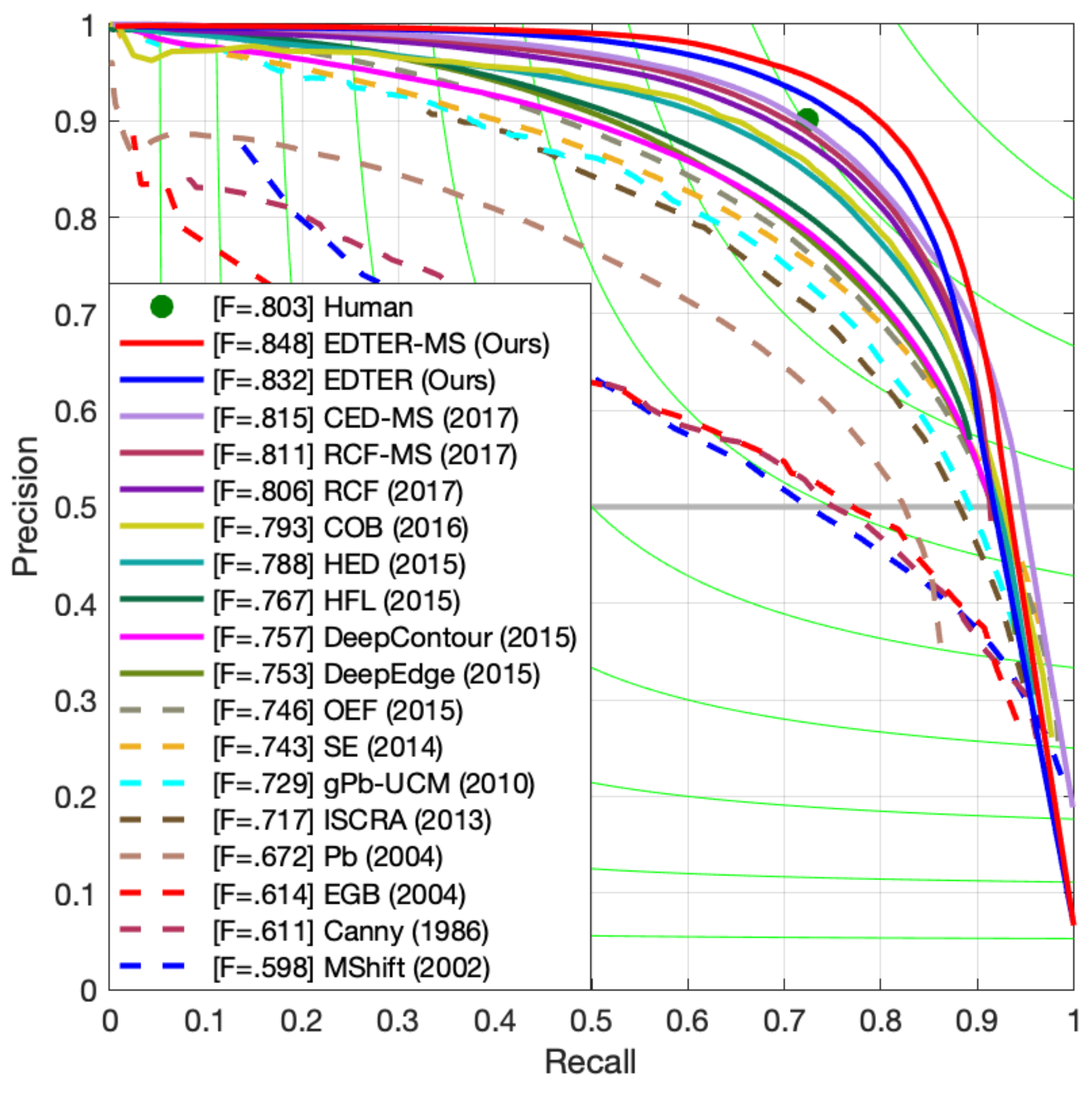}
\caption{The precision-recall curves on BSDS500.}
\label{fig:pr_bsds}
\vspace{-15pt}
\end{figure}

\begin{figure*}[!t]
\small
\setlength{\abovecaptionskip}{3pt}
\setlength{\belowcaptionskip}{0pt}
    \centering
	\begin{tabular}{cccccc}
		\hspace{-.2cm}
		\includegraphics[width=.155\textwidth]{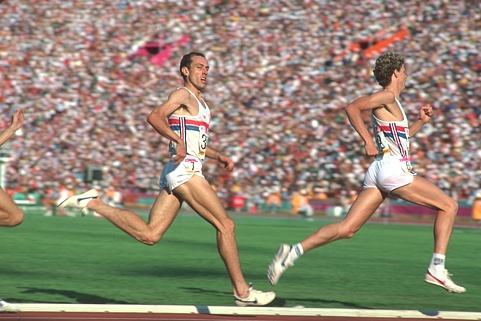} & \hspace{-.45cm}
		\includegraphics[width=.155\textwidth]{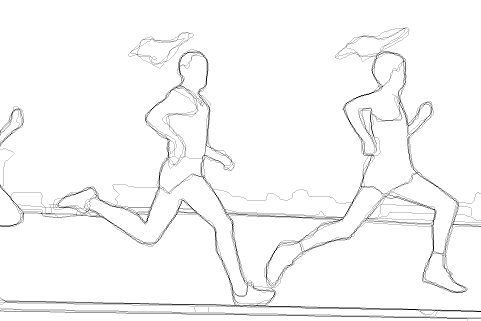} & \hspace{-.45cm}
		\includegraphics[width=.155\textwidth]{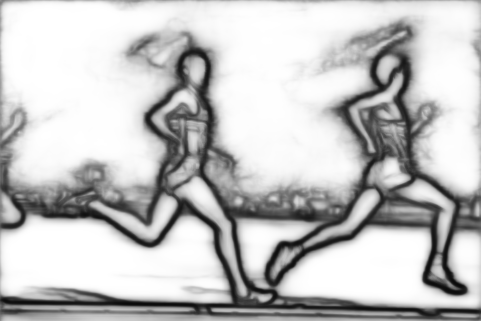} & \hspace{-.45cm}
		\includegraphics[width=.155\textwidth]{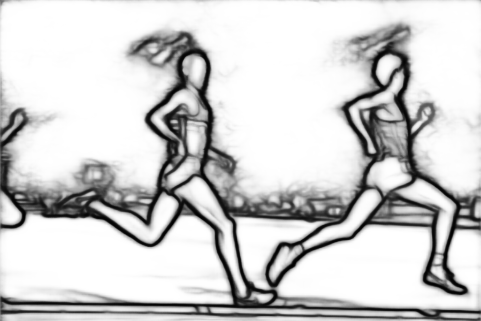} & \hspace{-.45cm}
		\includegraphics[width=.155\textwidth]{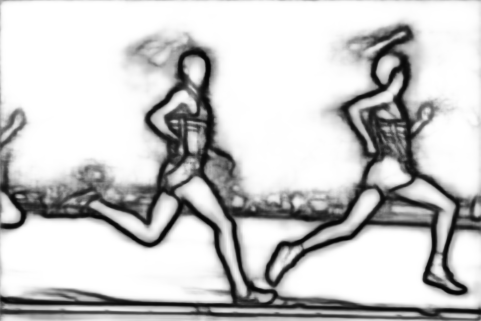} & \hspace{-.45cm}
		\includegraphics[width=.155\textwidth]{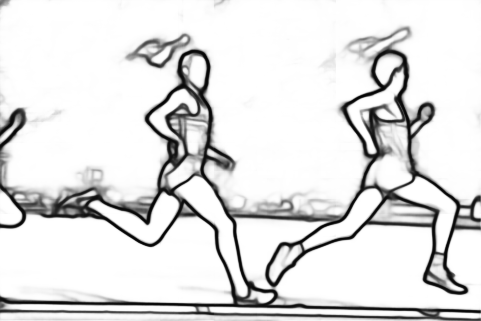}\vspace{-.06cm} \\
		\hspace{-.2cm}
		\includegraphics[width=.155\textwidth]{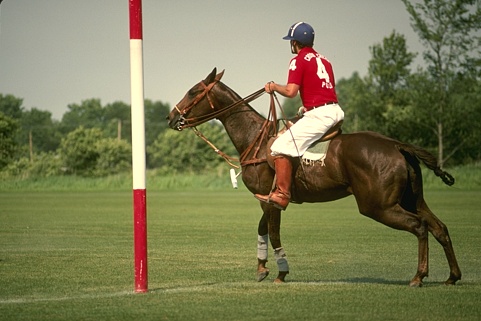} & \hspace{-.45cm}
		\includegraphics[width=.155\textwidth]{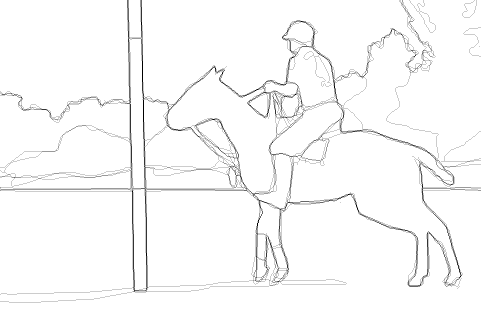} & \hspace{-.45cm}
		\includegraphics[width=.155\textwidth]{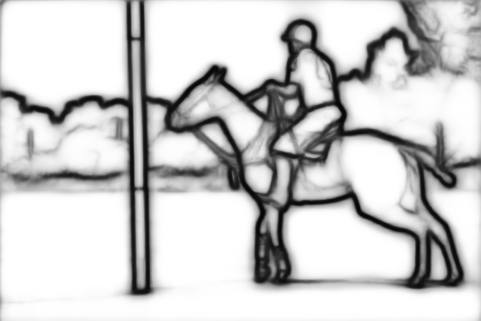} & \hspace{-.45cm}
		\includegraphics[width=.155\textwidth]{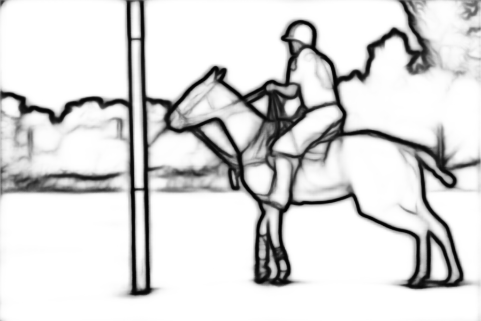} & \hspace{-.45cm}
		\includegraphics[width=.155\textwidth]{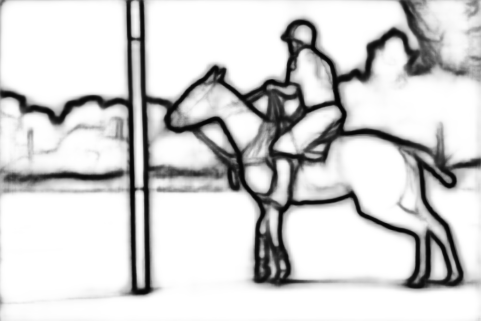} & \hspace{-.45cm}
		\includegraphics[width=.155\textwidth]{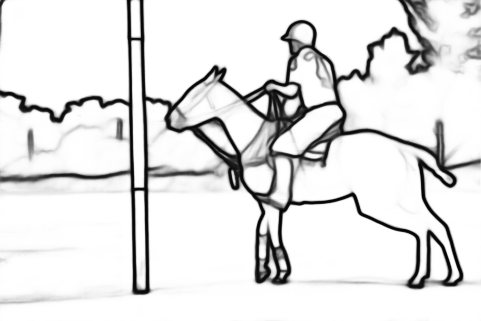}\vspace{-.06cm} \\
		\hspace{-.2cm}
		\includegraphics[width=.155\textwidth]{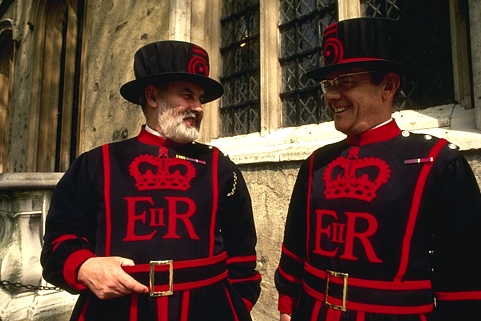} & \hspace{-.45cm}
		\includegraphics[width=.155\textwidth]{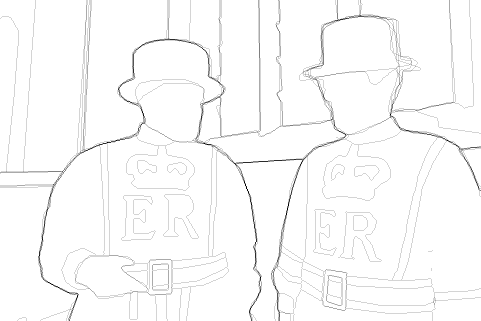} & \hspace{-.45cm}
		\includegraphics[width=.155\textwidth]{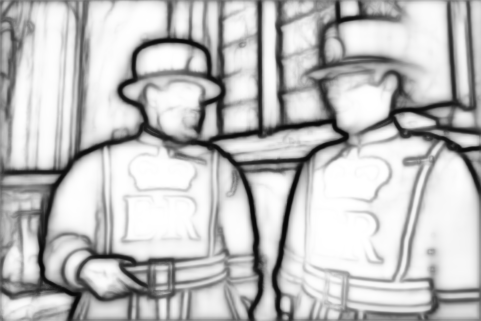} & \hspace{-.45cm}
		\includegraphics[width=.155\textwidth]{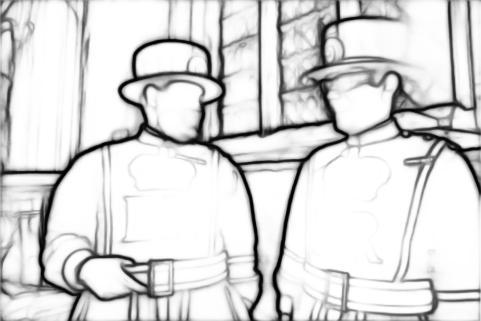} & \hspace{-.45cm}
		\includegraphics[width=.155\textwidth]{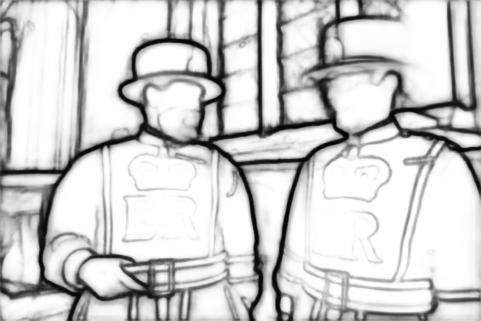} & \hspace{-.45cm}
		\includegraphics[width=.155\textwidth]{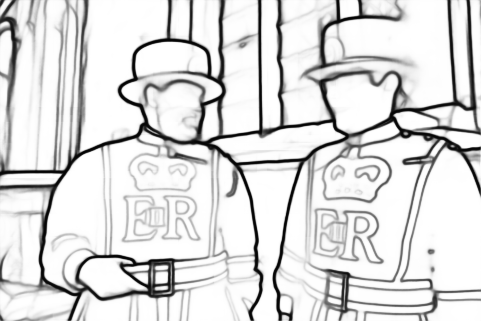}\vspace{-.06cm} \\
		\hspace{-.2cm} (a) Input 
		& \hspace{-.45cm} (b) Ground truth 
		&\hspace{-.45cm} (c) RCF~\cite{liu2017rcf} 
		&\hspace{-.45cm} (d) CED~\cite{wang2017ced} 
		&\hspace{-.5cm} (e) BDCN~\cite{he2019bdcn} &\hspace{-.45cm} (f) EDTER (Ours) \\
	\end{tabular}
    \caption{Qualitative comparisons on three challenging samples in the testing set of BSDS500.}
    \label{fig:fig_sota}
    \vspace{-12pt}
\end{figure*}

Quantitative results are shown in Table~\ref{tab:bsds}, and Fig.~\ref{fig:pr_bsds} shows Precision-Recall curves of all methods. By training on the trainval set of BSDS500, our method achieves the F-measure ODS of 0.824 with single-scale testing and obtains 0.840 with multi-scale inputs, which already outperforms most edge detectors. With the extra training data and multi-scale testing (following the settings of RCF, CED, BDCN, \etc), our method achieves 84.8\% (ODS), 86.5\% (OIS) 90.3\% (AP), which is superior to all the state-of-the-art edge detectors. Some qualitative results are shown in Fig.~\ref{fig:fig_sota}. We observe that the proposed EDTER shows a clear advantage in prediction quality, both crisp and accurate.

\vspace{0.2em}
\noindent{\bf On NYUDv2.} In NYUDv2, we conduct experiments on RGB images and compare against the state-of-the-art methods including gPb-ucm~\cite{arbelaez2010bsds}, Silberman~\etal\cite{silberman2012indoor}, gPb+NG~\cite{gupta2013perceptual}, SE~\cite{dollar2014fast}, SE+NG+~\cite{gupta2014seng}, OEF~\cite{hallman2015oef}, SemiContour~\cite{zhang2016semicontour}, HED~\cite{xie2015hed}, RCF~\cite{liu2017rcf}, AMH-Net~\cite{xu2017AMHNet}, LPCB~\cite{deng2018lpcb}, BDCN~\cite{he2019bdcn}, and PiDiNet~\cite{Su2021pidiNet}. All results are based on single-scale input. Table~\ref{tab:nyud} shows the quantitative results of our method and other competitors. Our method achieves the best scores of 77.4\%, 78.9\%, and 79.7\% of ODS, OIS, and AP, respectively. Compared to the second best, we increase the scores by 2.6\%, 2.6\%, 2.7\% in three metrics, respectively. More results including HHA, RGB-HHA inputs, and visualizations are reported in the \href{https://github.com/MengyangPu/EDTER}{supplementary material}.

\begin{table}%[h]
\setlength{\abovecaptionskip}{0pt}
\caption{Quantitative comparisons on NYUDv2~\cite{silberman2012indoor}. All results are computed with a single scale input.}
\centering
\footnotesize
\renewcommand\arraystretch{0.9}
\renewcommand\tabcolsep{5.0pt}
\begin{tabular}{c|l|c|ccc}
    \toprule
    \multicolumn{2}{c|}{Method} &\makecell[c]{Pub.'Year}    & ODS   & OIS   & AP        \\
    \toprule
    \multirow{7}{*}{\rotatebox{90}{Traditional}}
    & gPb-ucm~\cite{arbelaez2010bsds}     & PAMI'11     & 0.632 & 0.661 & 0.562     \\
    & Silberman \etal~\cite{silberman2012indoor}  & ECCV'12     & 0.658 & 0.661 & -  \\
    & gPb+NG~\cite{gupta2013perceptual}   & CVPR'13     & 0.687 & 0.716 & 0.629     \\
    & SE~\cite{dollar2014fast}            & PAMI'14     & 0.695 & 0.708 & 0.679     \\
    & SE+NG+~\cite{gupta2014seng}         & ECCV'14     & 0.706 & 0.734 & 0.738     \\
    & OEF~\cite{hallman2015oef}           & CVPR'15     & 0.651 & 0.667 & -         \\
    & SemiContour~\cite{zhang2016semicontour}       & CVPR'16   & 0.680 & 0.700 & 0.690   \\
    \toprule
    \multirow{6}{*}{\rotatebox{90}{CNN-based}}
    & HED~\cite{xie2015hed}       & ICCV'15     & 0.720 & 0.734 & 0.734     \\
    & RCF~\cite{liu2017rcf}       & CVPR'17     & 0.729 & 0.742 & -         \\
    & AMH-Net~\cite{xu2017AMHNet} & NeurIPS'17  & 0.744 & 0.758 & 0.765     \\
    & LPCB~\cite{deng2018lpcb}    & ECCV'18     & 0.739 & 0.754 & -         \\
    & BDCN~\cite{he2019bdcn}      & CVPR'19     & \SBEST{0.748} & \SBEST{0.763} & \SBEST{0.770}     \\
    & PiDiNet~\cite{Su2021pidiNet}& ICCV'21     & 0.733 & 0.747 & -         \\
    \toprule
    & EDTER (Ours)                & -           & \BEST{0.774}  & \BEST{0.789}  & \BEST{0.797}\\    
    \bottomrule
  \end{tabular}
\label{tab:nyud}
\vspace{-14pt}
\end{table}

\begin{table}[t]
\setlength{\abovecaptionskip}{0pt}
\caption{Comparisons on Multicue~\cite{mely2016multicue}. All results are computed with a single scale input.}
\centering
\footnotesize
\renewcommand\arraystretch{0.9}
\renewcommand\tabcolsep{1.5pt}
\begin{tabular}{c|l|c|ccc}
    \toprule
    & Method                          & Pub.'Year & ODS           & OIS           & AP            \\  
    \toprule
    \multirow{8}{*}{\rotatebox{90}{Edge}}
    & Human~\cite{mely2016multicue}   & VR'16     & .750 (0.024) & -             & -             \\
    & Multicue~\cite{mely2016multicue}& VR'16     & .830 (0.002) & -             & -             \\
    & HED~\cite{xie2015hed}           & ICCV'15   & .851 (0.014) & .864 (0.011) & -             \\
    & RCF~\cite{liu2017rcf}           & CVPR'17   & .857 (0.004) & .862 (0.004) & -             \\
    & BDCN~\cite{he2019bdcn}          & CVPR'19   & \SBEST{.891 (0.001)} & \SBEST{.898 (0.002)} & \SBEST{.935(0.002)}  \\
    & DSCD~\cite{deng2020dscd}        & ACMMM'20  & .871 (0.007) & .876 (0.002) & -             \\
    & PiDiNet~\cite{Su2021pidiNet}    & ICCV'21   & .855 (0.007) & .860 (0.005) & -              \\
    & EDTER (Ours)                    & -         & \BEST{.894 (0.005)} & \BEST{.900 (0.003)} & \BEST{.944 (0.002)}     \\
    \toprule
    \multirow{8}{*}{\rotatebox{90}{Boundary}}
    & Human~\cite{mely2016multicue}   & VR'16     & .760 (0.017) & -             & -             \\
    & Multicue~\cite{mely2016multicue}& VR'16     & .720 (0.014) & -             & -             \\
    & HED~\cite{xie2015hed}           & ICCV'15   & .814 (0.011) & .822 (0.008) & .869 (0.015) \\
    & RCF~\cite{liu2017rcf}           & CVPR'17   & .817 (0.004) & .825 (0.005) & -             \\
    & BDCN~\cite{he2019bdcn}          & CVPR'19   & \SBEST{.836 (0.001)} & \SBEST{.846 (0.003)} & \SBEST{.893 (0.001)} \\
    & DSCD~\cite{deng2020dscd}        & ACMMM'20  & .828 (0.003) & .835 (0.004) & -             \\
    & PiDiNet~\cite{Su2021pidiNet}    & ICCV'21   & .818 (0.003) & .830 (0.005) & -             \\
    & EDTER (Ours)                    & -         & \BEST{.861 (0.003)} & \BEST{.870 (0.004)} & \BEST{.919 (0.003)}\\
    \bottomrule
  \end{tabular}
\label{tab:multicue}
\vspace{-12pt}
\end{table}

\vspace{0.1em}
\noindent{\bf On Multicue.} Multicue consists of two kinds of annotations, \ie, Multicue Edge and Multicue Boundary. For each type of annotations, we compare against the state-of-the-art methods including HED~\cite{xie2015hed}, RCF~\cite{liu2017rcf}, BDCN~\cite{he2019bdcn}, DSCD~\cite{deng2020dscd}, and PiDiNet~\cite{Su2021pidiNet}. We report the comparisons in Table~\ref{tab:multicue}, and the results show consistent performance. Our method produces competitive results on Multicue Edge. For Multicue Boundary, EDTER achieves the F-measure ODS of 86.1\%, higher than all other competitors. The visualization results are provided in the \href{https://github.com/MengyangPu/EDTER}{supplementary material}.

%------------------------------------------------------------------
\section{Conclusion and Limitation}
In this paper, we propose a novel two-stage edge detection framework, namely EDTER. By introducing the vision transformer, EDTER captures both coarse-grained global context and fine-grained local context in two stages. Moreover, it employs a novel Bi-directional Multi-Level Aggregation (BiMLA) decoder to explore high-resolution representations. Besides, a Feature Fusion Module (FFM) incorporates global and local contexts to predict the edge results. Experimental results illustrate that EDTER yields competitive results in comparison with state-of-the-arts. 

\vspace{0.5em} 
\noindent\textbf{Limitation.} The width of the edges extracted by EDTER occupies multiple pixels, which still has a gap with the ideal edge width. Without any post-processing, generating clear and thin edges is still a future direction to explore.

\vspace{0.6em}
\noindent\textbf{Acknowledgements.} 
This work is supported by Beijing Natural Science Foundation (M22022, L211015), Fundamental Research Funds for the Central Universities (2019JBZ104), and National Natural Science Foundation of China (61906013, 62106017).

%----------------------------------------------------------------
%%%%%%%%% REFERENCES
{\small
\bibliographystyle{ieee_fullname}
\bibliography{egbib}

\begin{thebibliography}{10}\itemsep=-1pt

\bibitem{arbelaez2010bsds}
Pablo Arbelaez, Michael Maire, Charless Fowlkes, and Jitendra Malik.
\newblock Contour detection and hierarchical image segmentation.
\newblock {\em IEEE Trans. Pattern Anal. Mach. Intell.}, 33(5):898--916, 2010.

\bibitem{arbelaez2010contour}
Pablo Arbelaez, Michael Maire, Charless Fowlkes, and Jitendra Malik.
\newblock Contour detection and hierarchical image segmentation.
\newblock {\em IEEE Trans. Pattern Anal. Mach. Intell.}, 33(5):898--916, 2010.

\bibitem{bertasius2015deepedge}
Gedas Bertasius, Jianbo Shi, and Lorenzo Torresani.
\newblock Deepedge: A multi-scale bifurcated deep network for top-down contour
  detection.
\newblock In {\em IEEE Conf. Comput. Vis. Pattern Recog.}, pages 4380--4389,
  2015.

\bibitem{bertasius2015hfl}
Gedas Bertasius, Jianbo Shi, and Lorenzo Torresani.
\newblock High-for-low and low-for-high: Efficient boundary detection from deep
  object features and its applications to high-level vision.
\newblock In {\em Int. Conf. Comput. Vis.}, pages 504--512, 2015.

\bibitem{caelles2017one}
Sergi Caelles, Kevis-Kokitsi Maninis, Jordi Pont-Tuset, Laura Leal-Taix{\'e},
  Daniel Cremers, and Luc Van~Gool.
\newblock One-shot video object segmentation.
\newblock In {\em IEEE Conf. Comput. Vis. Pattern Recog.}, pages 221--230,
  2017.

\bibitem{canny1986computational}
John~F. Canny.
\newblock A computational approach to edge detection.
\newblock {\em IEEE Trans. Pattern Anal. Mach. Intell.}, 8(6):679--698, 1986.

\bibitem{carion2020detr}
Nicolas Carion, Francisco Massa, Gabriel Synnaeve, Nicolas Usunier, Alexander
  Kirillov, and Sergey Zagoruyko.
\newblock End-to-end object detection with transformers.
\newblock In {\em Eur. Conf. Comput. Vis.}, pages 213--229. Springer, 2020.

\bibitem{chen2017deeplab}
Liang-Chieh Chen, George Papandreou, Iasonas Kokkinos, Kevin Murphy, and Alan~L
  Yuille.
\newblock Deeplab: Semantic image segmentation with deep convolutional nets,
  atrous convolution, and fully connected crfs.
\newblock {\em IEEE Trans. Pattern Anal. Mach. Intell.}, 40(4):834--848, 2017.

\bibitem{chen2021transformertracking}
Xin Chen, Bin Yan, Jiawen Zhu, Dong Wang, Xiaoyun Yang, and Huchuan Lu.
\newblock Transformer tracking.
\newblock In {\em IEEE Conf. Comput. Vis. Pattern Recog.}, pages 8126--8135,
  2021.

\bibitem{Dai_2021_CVPR_UPDETR}
Zhigang Dai, Bolun Cai, Yugeng Lin, and Junying Chen.
\newblock Up-detr: Unsupervised pre-training for object detection with
  transformers.
\newblock In {\em IEEE Conf. Comput. Vis. Pattern Recog.}, pages 1601--1610,
  2021.

\bibitem{deng2020dscd}
Ruoxi Deng and Shengjun Liu.
\newblock Deep structural contour detection.
\newblock In {\em ACM Int. Conf. Multimedia}, pages 304--312, 2020.

\bibitem{deng2018lpcb}
Ruoxi Deng, Chunhua Shen, Shengjun Liu, Huibing Wang, and Xinru Liu.
\newblock Learning to predict crisp boundaries.
\newblock In {\em Eur. Conf. Comput. Vis.}, pages 562--578, 2018.

\bibitem{devlin2019bert}
Jacob Devlin, Ming-Wei Chang, Kenton Lee, and Kristina Toutanova.
\newblock Bert: Pre-training of deep bidirectional transformers for language
  understanding.
\newblock {\em NAACL}, 2019.

\bibitem{dollar2006supervised}
Piotr Doll{\'{a}}r, Zhuowen Tu, and Serge~J. Belongie.
\newblock Supervised learning of edges and object boundaries.
\newblock In {\em IEEE Conf. Comput. Vis. Pattern Recog.}, volume~2, pages
  1964--1971, 2006.

\bibitem{dollar2014fast}
Piotr Doll{\'a}r and C~Lawrence Zitnick.
\newblock Fast edge detection using structured forests.
\newblock {\em IEEE Trans. Pattern Anal. Mach. Intell.}, 37(8):1558--1570,
  2014.

\bibitem{dosovitskiy2020image16x16}
Alexey Dosovitskiy, Lucas Beyer, Alexander Kolesnikov, Dirk Weissenborn,
  Xiaohua Zhai, Thomas Unterthiner, Mostafa Dehghani, Matthias Minderer, Georg
  Heigold, Sylvain Gelly, et~al.
\newblock An image is worth 16x16 words: Transformers for image recognition at
  scale.
\newblock {\em Int. Conf. Learn. Represent.}, 2020.

\bibitem{everingham2010pascal}
Mark Everingham, Luc Van~Gool, Christopher~KI Williams, John Winn, and Andrew
  Zisserman.
\newblock The pascal visual object classes (voc) challenge.
\newblock {\em Int. J. Comput. Vis.}, 88(2):303--338, 2010.

\bibitem{felzenszwalb2004felz}
Pedro~F Felzenszwalb and Daniel~P Huttenlocher.
\newblock Efficient graph-based image segmentation.
\newblock {\em Int. J. Comput. Vis.}, 59(2):167--181, 2004.

\bibitem{gupta2013perceptual}
Saurabh Gupta, Pablo Arbelaez, and Jitendra Malik.
\newblock Perceptual organization and recognition of indoor scenes from rgb-d
  images.
\newblock In {\em IEEE Conf. Comput. Vis. Pattern Recog.}, pages 564--571,
  2013.

\bibitem{gupta2014seng}
Saurabh Gupta, Ross Girshick, Pablo Arbel{\'a}ez, and Jitendra Malik.
\newblock Learning rich features from rgb-d images for object detection and
  segmentation.
\newblock In {\em Eur. Conf. Comput. Vis.}, pages 345--360. Springer, 2014.

\bibitem{hallman2015oef}
Sam Hallman and Charless~C Fowlkes.
\newblock Oriented edge forests for boundary detection.
\newblock In {\em IEEE Conf. Comput. Vis. Pattern Recog.}, pages 1732--1740,
  2015.

\bibitem{he2019bdcn}
Jianzhong He, Shiliang Zhang, Ming Yang, Yanhu Shan, and Tiejun Huang.
\newblock Bi-directional cascade network for perceptual edge detection.
\newblock In {\em IEEE Conf. Comput. Vis. Pattern Recog.}, pages 3828--3837,
  2019.

\bibitem{He2017MaskR}
Kaiming He, Georgia Gkioxari, Piotr Doll{\'a}r, and Ross~B. Girshick.
\newblock Mask r-cnn.
\newblock {\em Int. Conf. Comput. Vis.}, pages 2980--2988, 2017.

\bibitem{hwang2015cscnn}
Jyh-Jing Hwang and Tyng-Luh Liu.
\newblock Pixel-wise deep learning for contour detection.
\newblock {\em arXiv preprint arXiv:1504.01989}, 2015.

\bibitem{isola2014pmi}
Phillip Isola, Daniel Zoran, Dilip Krishnan, and Edward~H Adelson.
\newblock Crisp boundary detection using pointwise mutual information.
\newblock In {\em Eur. Conf. Comput. Vis.}, pages 799--814. Springer, 2014.

\bibitem{kelm2019rcn}
Andr{\'e}~Peter Kelm, Vijesh~Soorya Rao, and Udo Z{\"o}lzer.
\newblock Object contour and edge detection with refinecontournet.
\newblock In {\em International Conference on Computer Analysis of Images and
  Patterns}, pages 246--258. Springer, 2019.

\bibitem{Kim_2021_CVPR_HOTR}
Bumsoo Kim, Junhyun Lee, Jaewoo Kang, Eun-Sol Kim, and Hyunwoo~J. Kim.
\newblock Hotr: End-to-end human-object interaction detection with
  transformers.
\newblock In {\em IEEE Conf. Comput. Vis. Pattern Recog.}, pages 74--83, 2021.

\bibitem{kittler1983accuracy}
Josef Kittler.
\newblock On the accuracy of the sobel edge detector.
\newblock {\em {Image Vis. Comput.}}, 1(1):37--42, 1983.

\bibitem{kokkinos2015pushing}
Iasonas Kokkinos.
\newblock Pushing the boundaries of boundary detection using deep learning.
\newblock {\em Int. Conf. Learn. Represent.}, 2016.

\bibitem{lan2020albert}
Zhenzhong Lan, Mingda Chen, Sebastian Goodman, Kevin Gimpel, Piyush Sharma, and
  Radu Soricut.
\newblock Albert: A lite bert for self-supervised learning of language
  representations.
\newblock {\em Int. Conf. Learn. Represent.}, 2020.

\bibitem{Li_2021_CVPR_Pose}
Ke Li, Shijie Wang, Xiang Zhang, Yifan Xu, Weijian Xu, and Zhuowen Tu.
\newblock Pose recognition with cascade transformers.
\newblock In {\em IEEE Conf. Comput. Vis. Pattern Recog.}, pages 1944--1953,
  2021.

\bibitem{Li_2021_Diverse}
Yulin Li, Jianfeng He, Tianzhu Zhang, Xiang Liu, Yongdong Zhang, and Feng Wu.
\newblock Diverse part discovery: Occluded person re-identification with
  part-aware transformer.
\newblock In {\em IEEE Conf. Comput. Vis. Pattern Recog.}, pages 2898--2907,
  2021.

\bibitem{liao2017dcd}
Yuan Liao, Songping Fu, Xiaoqing Lu, Chengcui Zhang, and Zhi Tang.
\newblock Deep-learning-based object-level contour detection with ccg and crf
  optimization.
\newblock In {\em Int. Conf. Multimedia and Expo}, pages 859--864. IEEE, 2017.

\bibitem{lim2013sketch}
Joseph~J. Lim, C.~Lawrence Zitnick, and Piotr Doll{\'{a}}r.
\newblock Sketch tokens: {A} learned mid-level representation for contour and
  object detection.
\newblock In {\em IEEE Conf. Comput. Vis. Pattern Recog.}, pages 3158--3165,
  2013.

\bibitem{LinFPN17}
Tsung{-}Yi Lin, Piotr Doll{\'{a}}r, Ross~B. Girshick, Kaiming He, Bharath
  Hariharan, and Serge~J. Belongie.
\newblock Feature pyramid networks for object detection.
\newblock In {\em IEEE Conf. Comput. Vis. Pattern Recog.}, pages 936--944,
  2017.

\bibitem{liu2017rcf}
Yun Liu, Ming-Ming Cheng, Xiaowei Hu, Kai Wang, and Xiang Bai.
\newblock Richer convolutional features for edge detection.
\newblock In {\em IEEE Conf. Comput. Vis. Pattern Recog.}, pages 3000--3009,
  2017.

\bibitem{liu2016rds}
Yu Liu and Michael~S Lew.
\newblock Learning relaxed deep supervision for better edge detection.
\newblock In {\em IEEE Conf. Comput. Vis. Pattern Recog.}, pages 231--240,
  2016.

\bibitem{liu2021swin}
Ze Liu, Yutong Lin, Yue Cao, Han Hu, Yixuan Wei, Zheng Zhang, Stephen Lin, and
  Baining Guo.
\newblock Swin transformer: Hierarchical vision transformer using shifted
  windows.
\newblock In {\em Int. Conf. Comput. Vis.}, 2021.

\bibitem{long2015fully}
Jonathan Long, Evan Shelhamer, and Trevor Darrell.
\newblock Fully convolutional networks for semantic segmentation.
\newblock In {\em IEEE Conf. Comput. Vis. Pattern Recog.}, pages 3431--3440,
  2015.

\bibitem{maninis2016cob}
Kevis-Kokitsi Maninis, Jordi Pont-Tuset, Pablo Arbel{\'a}ez, and Luc Van~Gool.
\newblock Convolutional oriented boundaries.
\newblock In {\em Eur. Conf. Comput. Vis.}, pages 580--596. Springer, 2016.

\bibitem{martin2004learning}
David~R. Martin, Charless~C. Fowlkes, and Jitendra Malik.
\newblock Learning to detect natural image boundaries using local brightness,
  color, and texture cues.
\newblock {\em IEEE Trans. Pattern Anal. Mach. Intell.}, 26(5):530--549, 2004.

\bibitem{mely2016multicue}
David~A M{\'e}ly, Junkyung Kim, Mason McGill, Yuliang Guo, and Thomas Serre.
\newblock A systematic comparison between visual cues for boundary detection.
\newblock {\em Vis. Res.}, 120:93--107, 2016.

\bibitem{paszke2017automatic}
Adam Paszke, Sam Gross, Soumith Chintala, Gregory Chanan, Edward Yang, Zachary
  DeVito, Zeming Lin, Alban Desmaison, Luca Antiga, and Adam Lerer.
\newblock Automatic differentiation in pytorch.
\newblock In {\em Adv. Neural Inform. Process. Syst.}, 2017.

\bibitem{pinheiro2015learning}
Pedro~OO Pinheiro, Ronan Collobert, and Piotr Doll{\'a}r.
\newblock Learning to segment object candidates.
\newblock In {\em Adv. Neural Inform. Process. Syst.}, pages 1990--1998, 2015.

\bibitem{pinheiro2016learning}
Pedro~O Pinheiro, Tsung-Yi Lin, Ronan Collobert, and Piotr Doll{\'a}r.
\newblock Learning to refine object segments.
\newblock In {\em Eur. Conf. Comput. Vis.}, pages 75--91, 2016.

\bibitem{PuHGL21iccv}
Mengyang Pu, Yaping Huang, Qingji Guan, and Haibin Ling.
\newblock Rindnet: Edge detection for discontinuity in reflectance,
  illumination, normal and depth.
\newblock In {\em Int. Conf. Comput. Vis.}, pages 6879--6888, October 2021.

\bibitem{ronneberger2015u}
Olaf Ronneberger, Philipp Fischer, and Thomas Brox.
\newblock U-net: Convolutional networks for biomedical image segmentation.
\newblock In {\em International Conference on Medical Image Computing and
  Computer-Assisted Intervention (MICCAI)}, pages 234--241, 2015.

\bibitem{shen2015deepcontour}
Wei Shen, Xinggang Wang, Yan Wang, Xiang Bai, and Zhijiang Zhang.
\newblock Deepcontour: A deep convolutional feature learned by positive-sharing
  loss for contour detection.
\newblock In {\em IEEE Conf. Comput. Vis. Pattern Recog.}, pages 3982--3991,
  2015.

\bibitem{silberman2012indoor}
Nathan Silberman, Derek Hoiem, Pushmeet Kohli, and Rob Fergus.
\newblock Indoor segmentation and support inference from rgbd images.
\newblock In {\em Eur. Conf. Comput. Vis.}, pages 746--760, 2012.

\bibitem{sironi2015mes}
Amos Sironi, Vincent Lepetit, and Pascal Fua.
\newblock Projection onto the manifold of elongated structures for accurate
  extraction.
\newblock In {\em Int. Conf. Comput. Vis.}, pages 316--324, 2015.

\bibitem{sironi2015msc}
Amos Sironi, Engin T{\"u}retken, Vincent Lepetit, and Pascal Fua.
\newblock Multiscale centerline detection.
\newblock {\em IEEE Trans. Pattern Anal. Mach. Intell.}, 38(7):1327--1341,
  2015.

\bibitem{poma2020dexined}
Xavier Soria, Edgar Riba, and {\'{A}}ngel~D. Sappa.
\newblock Dense extreme inception network: Towards a robust cnn model for edge
  detection.
\newblock In {\em IEEE Winter Conf. Appl. Comput. Vis.}, pages 1923--1932,
  2020.

\bibitem{Su2021pidiNet}
Zhuo Su, Wenzhe Liu, Zitong Yu, Dewen Hu, Qing Liao, Qi Tian, Matti
  Pietikainen, and Li Liu.
\newblock Pixel difference networks for efficient edge detection.
\newblock In {\em Int. Conf. Comput. Vis.}, pages 5117--5127, October 2021.

\bibitem{Tao2020hierarchical}
Andrew Tao, Karan Sapra, and Bryan Catanzaro.
\newblock Hierarchical multi-scale attention for semantic segmentation.
\newblock {\em arXiv preprint arXiv:2005.10821}, 2020.

\bibitem{vaswani2021scaling}
Ashish Vaswani, Prajit Ramachandran, Aravind Srinivas, Niki Parmar, Blake
  Hechtman, and Jonathon Shlens.
\newblock Scaling local self-attention for parameter efficient visual
  backbones.
\newblock In {\em IEEE Conf. Comput. Vis. Pattern Recog.}, pages 12894--12904,
  2021.

\bibitem{vaswani2017attention}
Ashish Vaswani, Noam Shazeer, Niki Parmar, Jakob Uszkoreit, Llion Jones,
  Aidan~N Gomez, {\L}ukasz Kaiser, and Illia Polosukhin.
\newblock Attention is all you need.
\newblock In {\em Adv. Neural Inform. Process. Syst.}, pages 5998--6008, 2017.

\bibitem{voigtlaender2019feelvos}
Paul Voigtlaender, Yuning Chai, Florian Schroff, Hartwig Adam, Bastian Leibe,
  and Liang-Chieh Chen.
\newblock Feelvos: Fast end-to-end embedding learning for video object
  segmentation.
\newblock In {\em IEEE Conf. Comput. Vis. Pattern Recog.}, pages 9481--9490,
  2019.

\bibitem{Wang_2021_CVPR_Tracker}
Ning Wang, Wengang Zhou, Jie Wang, and Houqiang Li.
\newblock Transformer meets tracker: Exploiting temporal context for robust
  visual tracking.
\newblock In {\em IEEE Conf. Comput. Vis. Pattern Recog.}, pages 1571--1580,
  2021.

\bibitem{wang2019fast}
Qiang Wang, Li Zhang, Luca Bertinetto, Weiming Hu, and Philip~HS Torr.
\newblock Fast online object tracking and segmentation: A unifying approach.
\newblock In {\em IEEE Conf. Comput. Vis. Pattern Recog.}, pages 1328--1338,
  2019.

\bibitem{wang2018sft}
Xintao Wang, Ke Yu, Chao Dong, and Chen Change~Loy.
\newblock Recovering realistic texture in image super-resolution by deep
  spatial feature transform.
\newblock In {\em IEEE Conf. Comput. Vis. Pattern Recog.}, pages 606--615,
  2018.

\bibitem{wang2021end}
Yuqing Wang, Zhaoliang Xu, Xinlong Wang, Chunhua Shen, Baoshan Cheng, Hao Shen,
  and Huaxia Xia.
\newblock End-to-end video instance segmentation with transformers.
\newblock In {\em IEEE Conf. Comput. Vis. Pattern Recog.}, pages 8741--8750,
  2021.

\bibitem{wang2017ced}
Yupei Wang, Xin Zhao, and Kaiqi Huang.
\newblock Deep crisp boundaries.
\newblock In {\em IEEE Conf. Comput. Vis. Pattern Recog.}, pages 3892--3900,
  2017.

\bibitem{winnemoller2011xdog}
Holger Winnem{\"{o}}ller, Jan~Eric Kyprianidis, and Sven~C. Olsen.
\newblock Xdog: An extended difference-of-gaussians compendium including
  advanced image stylization.
\newblock {\em Comput. Graph.}, 36(6):740--753, 2012.

\bibitem{xiaofeng2012scg}
Ren Xiaofeng and Liefeng Bo.
\newblock Discriminatively trained sparse code gradients for contour detection.
\newblock {\em Adv. Neural Inform. Process. Syst.}, 25, 2012.

\bibitem{xie2015hed}
Saining Xie and Zhuowen Tu.
\newblock Holistically-nested edge detection.
\newblock In {\em Int. Conf. Comput. Vis.}, pages 1395--1403, 2015.

\bibitem{xu2017AMHNet}
Dan Xu, Wanli Ouyang, Xavier Alameda-Pineda, Elisa Ricci, Xiaogang Wang, and
  Nicu Sebe.
\newblock Learning deep structured multi-scale features using attention-gated
  crfs for contour prediction.
\newblock In {\em Adv. Neural Inform. Process. Syst.}, pages 3961--3970, 2017.

\bibitem{yang2016cedn}
Jimei Yang, Brian Price, Scott Cohen, Honglak Lee, and Ming-Hsuan Yang.
\newblock Object contour detection with a fully convolutional encoder-decoder
  network.
\newblock In {\em IEEE Conf. Comput. Vis. Pattern Recog.}, pages 193--202,
  2016.

\bibitem{zhang2021multi}
Pengchuan Zhang, Xiyang Dai, Jianwei Yang, Bin Xiao, Lu Yuan, Lei Zhang, and
  Jianfeng Gao.
\newblock Multi-scale vision longformer: A new vision transformer for
  high-resolution image encoding.
\newblock {\em arXiv preprint arXiv:2103.15358}, 2021.

\bibitem{zhang2016semicontour}
Zizhao Zhang, Fuyong Xing, Xiaoshuang Shi, and Lin Yang.
\newblock Semicontour: A semi-supervised learning approach for contour
  detection.
\newblock In {\em IEEE Conf. Comput. Vis. Pattern Recog.}, pages 251--259,
  2016.

\bibitem{ZhaoPSPN17}
Hengshuang Zhao, Jianping Shi, Xiaojuan Qi, Xiaogang Wang, and Jiaya Jia.
\newblock Pyramid scene parsing network.
\newblock In {\em IEEE Conf. Comput. Vis. Pattern Recog.}, pages 6230--6239,
  2017.

\bibitem{Zhao17PyramidScene}
Hengshuang Zhao, Jianping Shi, Xiaojuan Qi, Xiaogang Wang, and Jiaya Jia.
\newblock Pyramid scene parsing network.
\newblock In {\em IEEE Conf. Comput. Vis. Pattern Recog.}, pages 6230--6239,
  2017.

\bibitem{zheng2021rethinking}
Sixiao Zheng, Jiachen Lu, Hengshuang Zhao, Xiatian Zhu, Zekun Luo, Yabiao Wang,
  Yanwei Fu, Jianfeng Feng, Tao Xiang, Philip~HS Torr, et~al.
\newblock Rethinking semantic segmentation from a sequence-to-sequence
  perspective with transformers.
\newblock In {\em IEEE Conf. Comput. Vis. Pattern Recog.}, pages 6881--6890,
  2021.

\bibitem{Zou_2021_CVPR_HOI}
Cheng Zou, Bohan Wang, Yue Hu, Junqi Liu, Qian Wu, Yu Zhao, Boxun Li, Chenguang
  Zhang, Chi Zhang, Yichen Wei, and Jian Sun.
\newblock End-to-end human object interaction detection with hoi transformer.
\newblock In {\em IEEE Conf. Comput. Vis. Pattern Recog.}, pages 11825--11834,
  2021.

\end{thebibliography}
}

\end{document}